\documentclass{article} 
\usepackage[final]{template}

\usepackage{amsmath,amsfonts,bm}

\def\eqref#1{equation~\ref{#1}}

\def\1{\bm{1}}

\DeclareMathAlphabet{\mathsfit}{\encodingdefault}{\sfdefault}{m}{sl}
\SetMathAlphabet{\mathsfit}{bold}{\encodingdefault}{\sfdefault}{bx}{n}

\usepackage{url}
\usepackage{pifont}
\usepackage{booktabs}
\usepackage{enumitem}
\usepackage{titletoc}
\usepackage{graphicx} 
\usepackage{booktabs} 
\usepackage{makecell} 
\usepackage{multirow}
\usepackage{algorithm}
\usepackage{algorithmicx}
\usepackage{algpseudocode}
\usepackage{subcaption} %

\makeatletter
\def\@trackname{}
\makeatother

\usepackage{tcolorbox}
\usepackage{listings}
\usepackage{xcolor}

\tcbuselibrary{listings, skins, breakable}

\definecolor{codebg}{RGB}{245, 245, 245}
\definecolor{kwblue}{RGB}{0, 0, 180}
\definecolor{strred}{RGB}{163, 21, 21}
\definecolor{defaultblack}{RGB}{40, 40, 40}
\definecolor{rred}{RGB}{255, 19, 49}
\lstdefinestyle{pythonstyle}{
  backgroundcolor=\color{codebg},
  basicstyle=\ttfamily\small\color{defaultblack}\bfseries,
  keywordstyle=\color{kwblue}\bfseries,
  stringstyle=\color{rred}\bfseries,
  commentstyle=\color{rred}\bfseries, 
  showstringspaces=false,
  breaklines=true,
  numbers=none
}

\newtcblisting{PyPrompt}[1]{%
  listing only,
  listing options={language=Python, style=pythonstyle},
  colback=codebg,
  colframe=black,
  breakable,
  enhanced,
  rounded corners, %
  fonttitle=\bfseries,
  coltitle=white,
  attach boxed title to top border={yshift=-1pt}, %
  boxed title style={colback=black, colframe=black, rounded corners}, %
  title={#1}
}

\tcbset{
  simpleprompt/.style={
    breakable, enhanced,
    boxrule=0.7pt, arc=2mm,
    colframe=black!80,              
    left=6pt,right=6pt,top=6pt,bottom=6pt,
    fonttitle=\bfseries,
  }
}
\newtcolorbox{prompt}[2][]{simpleprompt,
  colback=citecolor!6!white,        
  title={#2}, #1}

\definecolor{citecolor}{HTML}{0071bc}
\usepackage[pagebackref=false,breaklinks=true,letterpaper=true,colorlinks,citecolor=citecolor,bookmarks=false]{hyperref}
\usepackage{gradient-text}

\newcommand{\our}{VCode}

\newcommand{\metric}{CodeVQA}
\newcommand{\coder}{VCoder}

\newcommand{\newcolor}[1]{\gradientRGB{#1}{0,0,139}{0,191,255}}
\title{
\newcolor{\our}: a Multimodal Coding Benchmark with\\\newcolor{SVG as Symbolic Visual Representation}
}

\author{\textsuperscript{$1$}Kevin Qinghong Lin\thanks{Equal contribution.}\quad
\textsuperscript{$2$}Yuhao Zheng\footnotemark[1]\quad 
\textsuperscript{$3$}Hangyu Ran\footnotemark[1]\quad
\textsuperscript{$3$}Dantong Zhu\quad\\
\textbf{\textsuperscript{$3$}Dongxing Mao\quad
\textsuperscript{$4$}Linjie Li\quad
\textsuperscript{$1$}Philip Torr\quad
\textsuperscript{$3$}Alex Jinpeng Wang}
\\[4px]
\textsuperscript{$1$}University of Oxford\quad
\textsuperscript{$2$}University of Science and Technology of China\quad
\\
\textsuperscript{$3$}Central South University\quad
\textsuperscript{$4$}Microsoft Research
\\[4px]
Project Page: \url{https://csu-jpg.github.io/VCode}
}

\newcommand{\eg}{\textit{e.g.,}}
\newcommand{\ie}{\textit{i.e.,}}

\begin{document}

\maketitle

\begin{abstract}
Code has emerged as a precise, executable medium for reasoning and action in the agent era. Yet progress has largely focused on linguistic-centric tasks (\eg~program synthesis, debugging), leaving \textit{visual-centric coding} underexplored. Conventional image representations rely on dense RGB pixels that capture appearance but provide limited symbolic abstraction. Inspired by how humans reason over sketches, we advocate {SVG code} as a compact, interpretable, and executable visual representation.
We introduce \textbf{\our}, a benchmark that reframes multimodal understanding as code generation: given an image, a model must produce SVG that preserves symbolic meaning for downstream reasoning. \our~covers three challenging domains—general commonsense (MM-Vet), professional disciplines (MMMU), and visual-centric perception (CV-Bench). To assess symbolic fidelity, we propose \textbf{CodeVQA}, a novel evaluation protocol in which \textit{a policy model answers questions over rendered SVG; correct answers indicate faithful symbolic preservation}.
Empirically, frontier VLMs struggle to generate faithful SVGs, revealing a persistent gap between language-centric and visual-centric coding. To close this gap, we introduce \textbf{VCoder}, an agentic framework that augments VLMs along two axes: (i) \emph{Thinking with Revision}, which iteratively analyzes discrepancies and refines SVG code; and (ii) \emph{Acting with Visual Tools}, where detectors and parsers supply structured cues (objects, shapes, text) beyond intrinsic model capacity. 
Across benchmarks, frontier VLMs with strong reasoning score well overall yet remain limited on professional knowledge and 3D reasoning; \coder\ delivers a \({+12.3}\)-point overall gain over the top-performing Claude-4-Opus. Human studies show that both humans and VLMs perform worse on rendered SVGs, their consistency reveals the promise of symbolic visual representation
The benchmark and code is available at \url{https://github.com/CSU-JPG/VCode}.
\end{abstract}

\section{Introduction}
To advance reasoning and agentic intelligence, {code} has emerged as a powerful medium for interacting with digital environments \cite{generativeagents,voyager,agentbench}. Unlike natural language, which is free-form and descriptive, code is precise, structured, and executable—making it an effective mechanism for action. Consequently, recent benchmarks have predominantly emphasized {\textit{linguistic-centric}} coding abilities, covering tasks such as program synthesis, debugging, and competitive programming \cite{humaneval,mbpp,livecodebench_v6,scicode,swebench}, where success is measured by both correctness and executability.
In the multi-modal regime, coding plays a crucial role in generating executable programs that interface with tools or environments to accomplish complex task, a paradigm that has gained particular traction in embodied agents \cite{voyager,code_as_policies,videogui,showui}.
A parallel line of work leverages code to generate \textit{synthetic} visual assets—such as charts \cite{chartmimic, plot2code}, diagrams \cite{starvector, svggenius}, or websites \cite{pix2code, design2code}—which synthesis assets, are not directly grounded in the natural visual world.

\begin{figure}[!t]
\centering
\includegraphics[width=\textwidth]{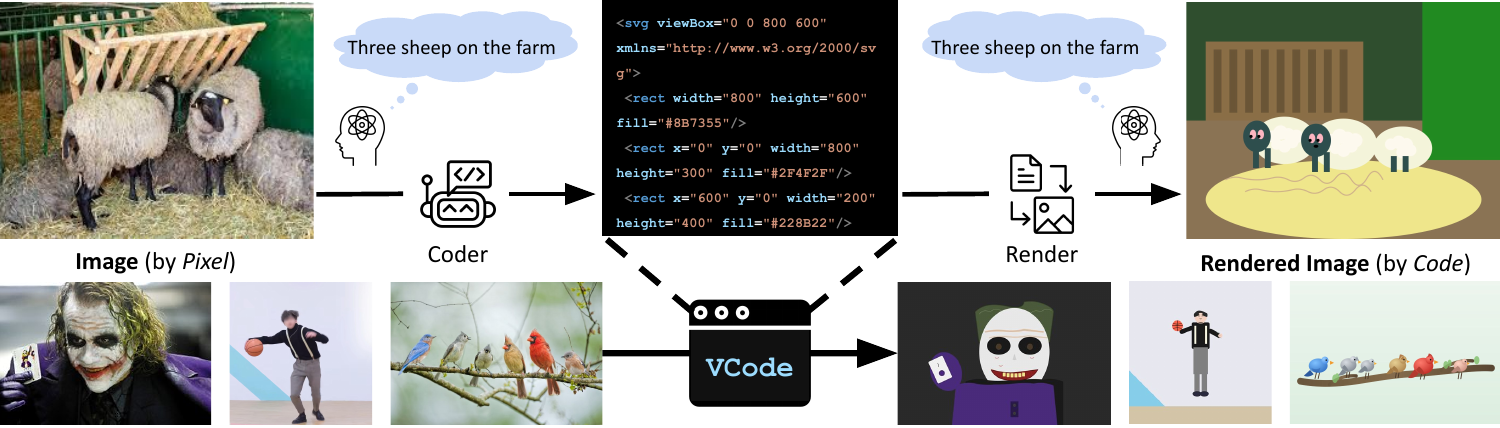}
\caption{\small{\textbf{Illustration of VCode.} An RGB image (left, represented by pixels) is translated into symbolic SVG code (middle) via VLM as Coder and rendered back into an image (right, represented by code), aiming to preserve symbolic meaning (\eg~“three sheep on the farm”). 
As shown at the bottom, VCode provides a compact, interpretable, and executable representation of original images.
\vspace{-1.5em}
}}
\label{fig:teaser}
\end{figure}

When recapping the representation for natural images, the dominant practice has been to encode them as pixels or superpixels. These representations are effective in that they densely capture visual apperance.
In contrast, humans often perceive and reason through sparse symbolic sketches that emphasize spatial relationships, object counts, and structural outlines \cite{visualsketchpad}. 
Similar to an artist drafting a rough sketch before filling in appearance details, such abstraction offers a compact yet informative scaffold for reasoning.
Building on this intuition, we propose using Scalable Vector Graphics (SVG) code as an alternative visual representation, owing to its compact, interpretable, and executable nature.
Thus, SVGs have long been used for icons and logos \cite{iconshop,starvector,omnisvg} for a general visual abstraction.
This perspective motivates a fundamental question: \textit{can visual representation move beyond raw pixels and learn to represent and reason through code?}

In this work, we introduce \our, a multimodal coding benchmark that pioneers the use of SVG code as a visual representation. \our~is constructed by repurposing existing multimodal understanding  benchmarks across three domains: General commonsense (MM-Vet~\cite{mmvet}), College-level disciplines (MMMU~\cite{mmmu}), and Visual-centric Perception (\eg~3D depth and relationships in CV-Bench~\cite{cvbench}).
\our~reframes these tasks as visual coding: given an image, a model must generate SVG code that faithfully renders the image, thereby reconstructing its symbolic representation. To evaluate this transformation, we propose \textbf{\metric}, a novel protocol in which a vision–language model must answer core questions about the original image by reasoning over the rendered SVG. This provides a principled test of \textit{whether the generated code serves as an adequate and faithful visual representation}.
Experiments on \our~show that existing coders remain limited in such challenging setting.
We observe that coding quality improves with a model’s reasoning ability, yet models still fail to preserve fine-grained visual relations (\eg~near vs. far), exposing a persistent gap between language- and visual-centric coding. 

To this end, we augment existing coders with two complementary capabilities. 
\textbf{(i) Thinking with Revision.} The model compares intermediate renderings with the original image, explicitly articulates discrepancies, and iteratively updates the SVG to improve fidelity. 
\textbf{(ii) Acting with Visual Tools.} We equip the coder with external perception toolboxes—\eg~object detectors and segmenters~\cite{florence2,sam2}—to supply structured cues (objects, shapes, text) as coding context. 
Together, these strategies yield a $+12.3$ overall gain over the top-performing Claude-4-Opus, substantially strengthening visual-centric coding.
Our contributions are threefold:

\begin{enumerate}[leftmargin=*, itemsep=0pt, topsep=0pt]
\item \textbf{\our: A Novel Perspective for Multimodal Coding.} We recast multimodal understanding as \emph{visual-centric coding}: given an image, generate SVG that preserves symbolic structure for downstream reasoning. We further present \textbf{\metric} -- a protocol that asks a VLM to answer the \emph{original-image} questions using only the \emph{rendered SVG}, thereby testing whether the code is an adequate and faithful visual representation.

  \item \textbf{\coder: Augmenting VLM as Strong Multimodal Coders} via \textit{(i) Thinking with Revision} (iterative discrepancy analysis and SVG refinement) and \textit{(ii) Acting with Visual Tools} (structured visual cues from detectors). \coder~achieves a significant overall gain over a strong baseline.

\item \textbf{Evaluation and Insights.} Extensive experiments expose persistent weaknesses of frontier VLMs in visual-centric coding. Human studies show consistent patterns between humans and VLMs when reasoning over rendered SVGs compared to raw images, suggesting the promise of symbolic visual coding for advancing human-like multimodal intelligence.
\end{enumerate}
\section{Related Works}
\subsection{Coding Benchmarks}
\begin{table*}[!t]
\centering
\scriptsize
\resizebox{\textwidth}{!}
{
\begin{tabular}{l l l l l l}
\toprule
\textbf{Benchmarks} & \textbf{Domain} & \textbf{Size} & \textbf{Inputs} & \textbf{Outputs} & \textbf{Evaluation}\\
\midrule
\multicolumn{5}{l}{\textcolor{gray}{\textit{Coding}}} \\
HumanEval \cite{humaneval} & Algorithm & 164 & Text & Code & Execute Pass\\
MMCode \cite{mmcode} & Visualization & 263 & Text \& Img & Code & Execute Pass \\
ChartMimic \cite{chartmimic} & Chart & 4800 & Text \& Img & Code & Similarity\\
Design2Code \cite{design2code} & Web UI & 484 & Text \& Img & Code & Similarity\\
SWE-Bench \cite{swebench} & GitHub & 2294 & Text \& Code & Code & Execute Pass \\
SVG-Bench \cite{starvector} & SVG & 23K & Img / Text & Code & Similarity \\ 
\midrule
\multicolumn{5}{l}{\textcolor{gray}{\textit{Multi-modal}}} \\
MM-Vet \cite{mmvet} & General & 218 & Img. \& text & Text & OpenQA\\
MMBench \cite{mmbench} & General & 3217 & Img. \& text & Text & MCQ\\
MMMU \cite{mmmu} & College & 11.5K & Img. \& text & Text & OpenQA / MCQ \\
MMMU-Pro \cite{mmmu-pro} & College & 1730 & Img. \& text & Text & OpenQA / MCQ \\
CV-Bench \cite{cvbench} & Perception & 2638 & Img. \& text & Text & MCQ\\
\textbf{\our~(Ours)} & \textbf{G\&C\&P} & 464 & Img. & Code & \textbf{Render$\rightarrow$VQA} \\
\bottomrule
\end{tabular}
}
\caption{
\small
\textbf{Comparison of \our\ with coding (top) and multimodal (bottom) benchmarks.}
\our\ differs in three ways:
\textbf{(i) {Task}}: models must generate \emph{code} directly from natural images, without extra query guidance;
\textbf{(ii) {Scope}}: focuses on natural multimodal understanding across diverse domains—General (G), College (C), and Perception (P);
\textbf{(iii) {Evaluation}}: introduces \textbf{\metric} (Render\,$\rightarrow$\,VQA), which judges whether the rendered SVG preserves the original image’s symbolic meaning.
}
\label{tab:benchmark}
\vspace{-2em}
\end{table*}

\textbf{Coding in LLMs.}
Despite there being several coding benchmarks, most of them are initially developed for purely language coding. Representative efforts include {HumanEval} \cite{humaneval} and {MBPP} \cite{mbpp}, which evaluate correctness of synthesized programs given natural language or code-level prompts. Later benchmarks such as {SWE-Bench} \cite{swebench} extend this paradigm to real-world software engineering, requiring models to resolve issues directly in large GitHub repositories. Despite their diversity, these benchmarks are fundamentally \textbf{linguistic-centric}: the inputs and outputs remain in textual or code form, with success measured by pass rates or test-case execution. While effective in quantifying reasoning over program text, such settings offer little insight into multimodal capabilities.

\textbf{Coding in Multi-modal.} Moving beyond purely textual code, a line of work incorporates {visual observations} into coding tasks. Benchmarks such as {Plot2Code} \cite{plot2code},{Design2Code} \cite{design2code}, and {ChartMimic} \cite{chartmimic} translate charts or UI mockups into executable plotting or layout code. {MMCode} \cite{mmcode} and {SWE-Bench-MM} \cite{swebenchmm} further integrate images alongside text, exploring how multimodal inputs can inform code generation. At larger scale, {SVGenius} \cite{svggenius} (generation, editing, understanding) evaluates models’ ability to produce vector-graphic code, highlighting challenges in preserving both semantics and structure.  
Despite this progress, most of these datasets emphasize \textbf{synthetic} visual assets (\eg~charts, Web UI, vector icons) as shown in Tab.\ref{tab:benchmark} top-half, leaving open the question of whether models can encode {real-world natural images} into executable visual code. This gap motivates our \our~benchmark, which repurposes multimodal understanding tasks into the visual coding with SVG.

\subsection{Multimodal Understanding}
Various benchmarks systematically evaluate multimodal understanding. Early efforts such as {MMBench}~\cite{mmbench} and {MM-Vet}~\cite{mmvet} emphasize general perception and text–image reasoning. More recent benchmarks, including {MMMU}~\cite{mmmu} and {MMMU-Pro}~\cite{mmmu-pro}, target professional knowledge and domain-specific reasoning. However, most of these evaluations interact with models through \textit{natural language} (\eg~query or answer).
In \our, we argue that generating \textit{code} to represent natural images constitutes an even more advanced form of understanding. 
As illustrated in Tab.\ref{tab:benchmark} bottom-half, unlike traditional perception tasks, this requires the model to distill an image into its {core concepts and structural features} by a \textit{render} image, and to express them in a symbolic format that bridges perception with reasoning and action. 
\section{\our~Benchmark}
\subsection{Task Definitions}
As illustrated in Fig.\ref{fig:teaser}, given an input RGB image $\mathcal{V}$, a vision–language model $\psi$ is tasked with generating SVG code $\mathcal{C}$ that encodes the image. 
Rendering this code yields a rendered image $\widetilde{\mathcal{V}}$. The objective is to minimize the discrepancy between the symbolic information of the original and rendered images:
\begin{equation}
\mathcal{L}=\min \big| I(\mathcal{V}) - I(\widetilde{\mathcal{V}}) \big| ,
\end{equation}
where $I(\cdot)$ denotes a symbolic information representation. The central challenge, however, lies in defining an applicable measure of symbolic information, which we elaborate on below.

\subsection{Evaluation Metrics}
The key to the evaluation prototype lies in how the correspondence between the input image $\mathcal{V}$ and the rendered image $\widetilde{\mathcal{V}}$ is defined.

\textbf{SigLIP score.} 
To define what constitutes an ideal SVG representation, we argue that it should faithfully preserve the semantic content of the original image rather than merely matching pixel-level similarity. One way to measure this is through embedding consistency. We leverage a pretrained visual encoder $f(\cdot)$ such as SigLIP~\cite{siglip,siglip2} to extract embeddings for both $\mathcal{V}$ and $\widetilde{\mathcal{V}}$, and compute their cosine distance:
\begin{equation}
\mathcal{L} = \max \cos\big( f(\mathcal{V}), f(\widetilde{\mathcal{V}}) \big).
\end{equation}

\textbf{\metric.} 
A more direct criterion is whether the rendered image $\widetilde{\mathcal{V}}$ alone supports correct reasoning.
Usually, $\widetilde{\mathcal{V}}$ may even facilitate answering questions that are ambiguous or harder to resolve from the original $\mathcal{V}$. Hence, the evaluation should not be constrained by the original image’s responses, but instead focus directly on the correctness of answers derived from $\widetilde{\mathcal{V}}$.
We define a \textit{policy} model $\phi$ that outputs an answer $\mathcal{A}$ given an image and a question $\mathcal{Q}$.
Then goal is formulated as
\begin{equation}
\begin{aligned}
\mathcal{A} &= \phi\!\left(\widetilde{\mathcal{V}}, \mathcal{Q}\right),\\
\mathcal{L} &= \max\mathbf{1}\!\left[\texttt{Evaluator}\!\left(\mathcal{A}\right)\right].
\end{aligned}
\end{equation}
where $\mathbf{1}[\cdot]$ is the indicator function. \texttt{Evaluator}$(\cdot)$ is a rule-based matching in multiple-choices setting, and it can be a LLM-as-Judge in open-ending.
If the answer is correct, the SVG suffices to convey the required semantics; otherwise, it reveals a gap in representational fidelity.

\textbf{Code tokens length.} Beyond faithful representation, we argue that an effective coder should represent an image with as few code tokens as possible, producing a concise yet faithful representation. To assess this efficiency, we evaluate the length of the generated SVG in terms of its token count $|\mathcal{C}|$.

\begin{figure}[!t]
    \centering
    \begin{subfigure}[t]{0.32\textwidth}
        \centering
        \includegraphics[width=\linewidth]{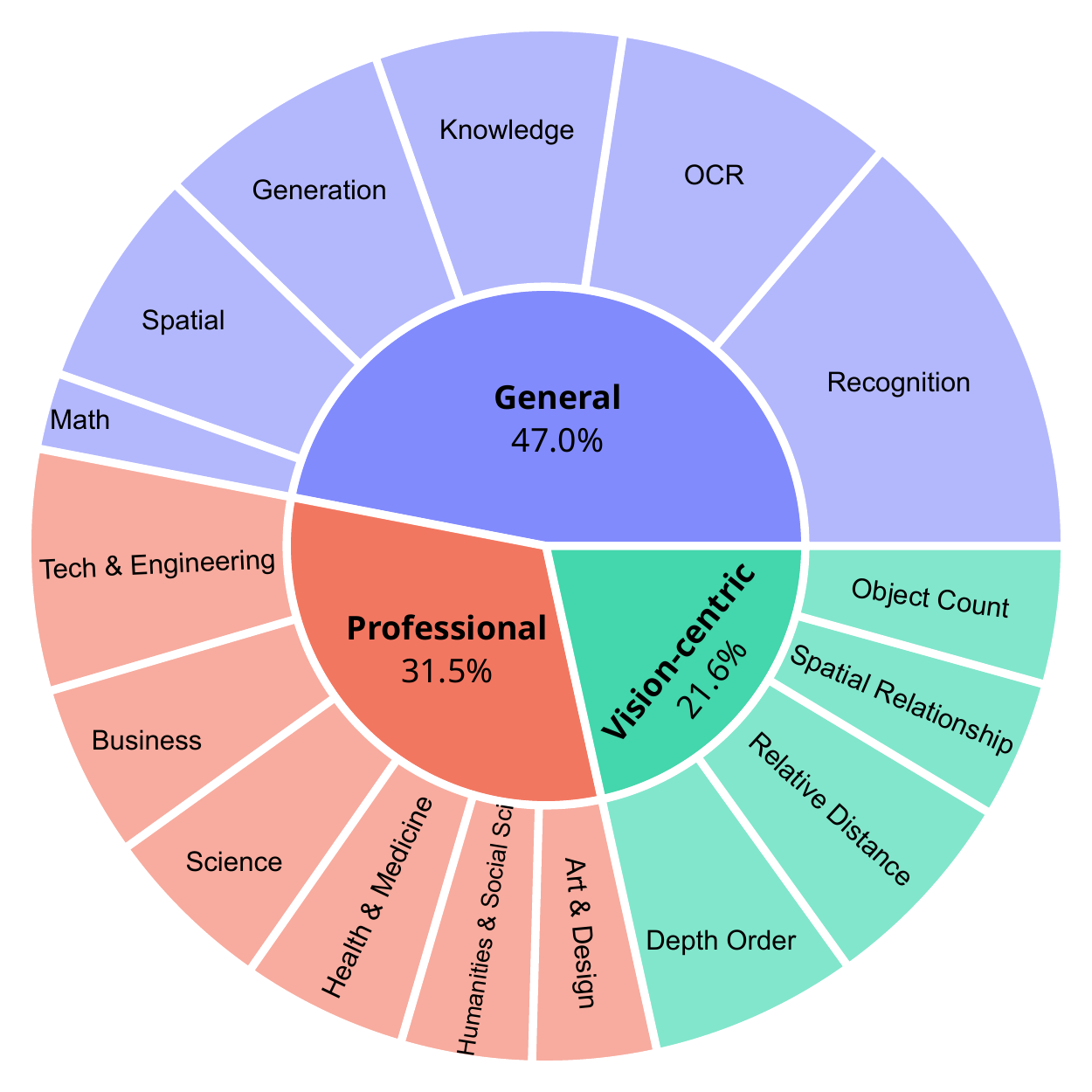}
        \caption{\textbf{Distributions of VCode.}}
    \end{subfigure}%
    \hfill
    \begin{subfigure}[t]{0.67\textwidth}
        \centering
        \includegraphics[width=\linewidth]{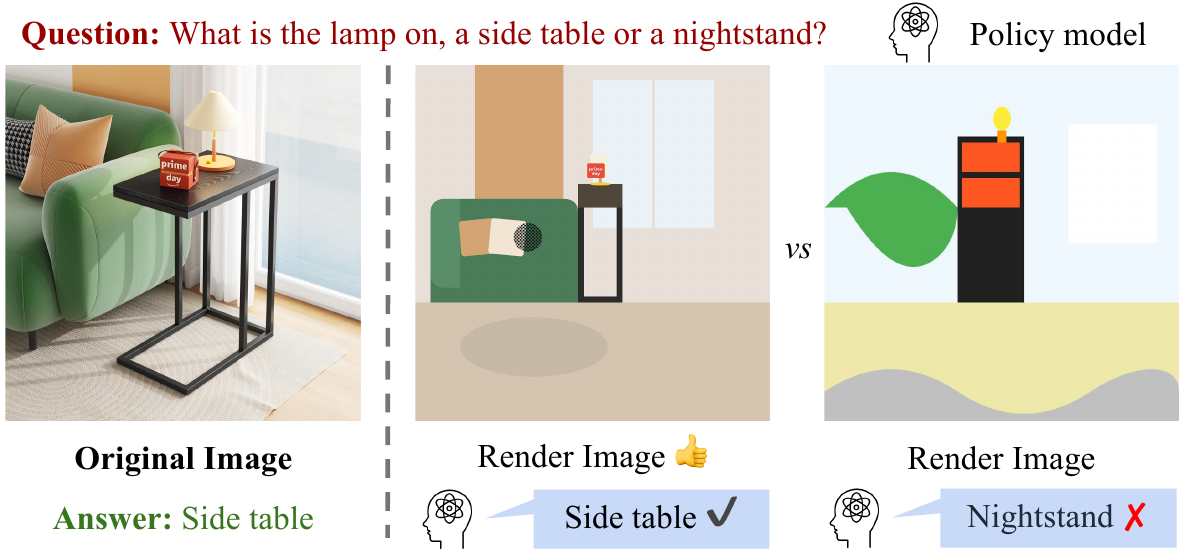}
        \caption{\textbf{Illustration of \metric~prototype.}}
    \end{subfigure}
\caption{\small{\textbf{Left: Distributions of tasks in VCode}, showing the proportions of general, professional, and vision-centric categories. \textbf{Right: Illustration of the \metric~prototype}: given an image and a question (\eg~“What is the lamp on, a side table or a nightstand?”), the policy model answers based on the rendered image. A correct answer indicates that the SVG representation preserves the semantic content of the original image, while an incorrect answer highlights room for improvement.}}
\vspace{-1.5em}
\label{fig:qa}
\end{figure}

\subsection{Data Curation}
With the evaluation prototype in place, the next step is to develop appropriate question sets $\mathcal{Q}$ for each associated image $\mathcal{V}$. To this end, we repurpose existing multimodal visual question answering benchmarks to align with our objective. To ensure diversity in taxonomy and difficulty, we focus on three representative scenarios:
\textbf{(i) Commonsense perception}: Assesses a model’s ability to capture everyday semantics such as spatial relationships. We adopt MM-Vet~\cite{mmvet} as the source.
\textbf{(ii) Professional knowledge}: Targets domain-specific, diploma-level tasks that demand both reasoning and coding skills. We incorporate the development set of MMMU~\cite{mmmu}, which spans multiple disciplines and requires deeper reasoning and expert knowledge.
\textbf{(iii) Visual-centric}: Evaluates performance in visually intensive settings involving counting, distance estimation, and relative spatial relationships in 2D or 3D. We draw from CV-Bench~\cite{cvbench}.

\textbf{Data statistics.}
Following this three-pronged curation strategy, we processed each source benchmark to construct our final dataset. For (i) commonsense perception, we incorporated the entirety of MM-Vet~\cite{mmvet}, resulting in 218 image-question pairs. For (ii) professional knowledge, our curation involved filtering the MMMU~\cite{mmmu} development set to retain only single-image VQA instances, which yielded a specialized subset of 146 pairs. Finally, for (iii) visual-centric, we created a balanced 100-pair subset from CV-Bench~\cite{cvbench} through a stratified sampling process. This involved shuffling the data and applying interval selection to ensure a specific distribution across its sub-tasks: spatial relationship (20), object count (20), depth order (30), and relative distance (30). In total, this process yields 464 image-question pairs.
The taxonomy distribution of \our~is illustrated in Fig.~\ref{fig:qa}(a).
\section{\coder}
In practice, we find that directly prompting Coders to generate SVG code from natural images remains highly challenging. This difficulty arises from three factors:
\textbf{(i) Long-Context Code Inputs:} fully representing an image typically requires thousands of tokens; composing such long sequences demands strong code reasoning over complex elements, beyond what current Coders provide.
\textbf{(ii) Visually-Blind Outputs:} inputs and outputs are cross-modal; because the rendered image is unseen until execution, the model must anticipate the visual consequences of code edits during generation.
\textbf{(iii) Weak Visual Fineness:} for irregular objects (\eg~a dog’s boundary), language models struggle to capture low-level details—edges, masks, and colors—that must be encoded precisely as numeric values, even though these are fundamental to code-based representations.

To address these intertwined challenges, we propose augmenting Coders at test time with two complementary capabilities.
\textbf{(a) Thinking with Revision:} we enhance reasoning through test-time scaling and a revision strategy that allows the model to iteratively refine its outputs, bridging the gap between long-context code generation and faithful visual rendering.
\textbf{(b) Acting with Vision Tools:} we equip Coders with external tools that extract fine-grained visual cues—such as edges, masks, and color regions—and translate them into structured code signals, enabling models to overcome their inherent limitations in low-level perception.

\subsection{Thinking with Revision}  
Since the initial reconstruction may not always yield a satisfactory result, a natural way to enhance Coders is to let them re-examine their own outputs and iteratively refine the code.  
Our revision strategy follows a two-step loop: detect discrepancies between the rendered output and the target image, then update the code conditioned on these differences.  

\textbf{(\textit{i}) Comment the Difference.}  
Given an intermediate rendering $\widetilde{\mathcal{V}}^{(t)}$, the coder first perceives its deviation from the original image $\mathcal{V}$. Although VLMs may be limited as Coders, they are already strong in visual perception. We therefore design the revision process to let them capture differences through two observations. At each iteration $t$, we compute a difference signal $\Delta^{(t)} \gets \psi\big(\mathcal{V}, \widetilde{\mathcal{V}}^{(t)}\big)$, which quantifies the discrepancy between the reconstruction and the target.  

\textbf{(\textit{ii}) Revise with Render Feedback.}  
The difference signal $\Delta^{(t)}$, together with the current code $\mathcal{C}^{(t)}$ and render $\widetilde{\mathcal{V}}^{(t)}$, is provided to the coder $\psi$ to generate revised code $\mathcal{C}^{(t+1)}$. Executing this code produces an updated reconstruction $\widetilde{\mathcal{V}}^{(t+1)} \leftarrow \big(\mathcal{V}, \widetilde{\mathcal{V}}^{(t)}, \mathcal{C}^{(t)}, \Delta^{(t)}\big)$.  

This revision loop is repeated for $t=0,1,\ldots,T$, progressively refining the reconstruction until a satisfactory visual outcome is reached. The full procedure is summarized in Algorithm~\ref{alg:iterative-revision}.

\begin{figure}[!t]
    \centering
    \includegraphics[width=\textwidth]{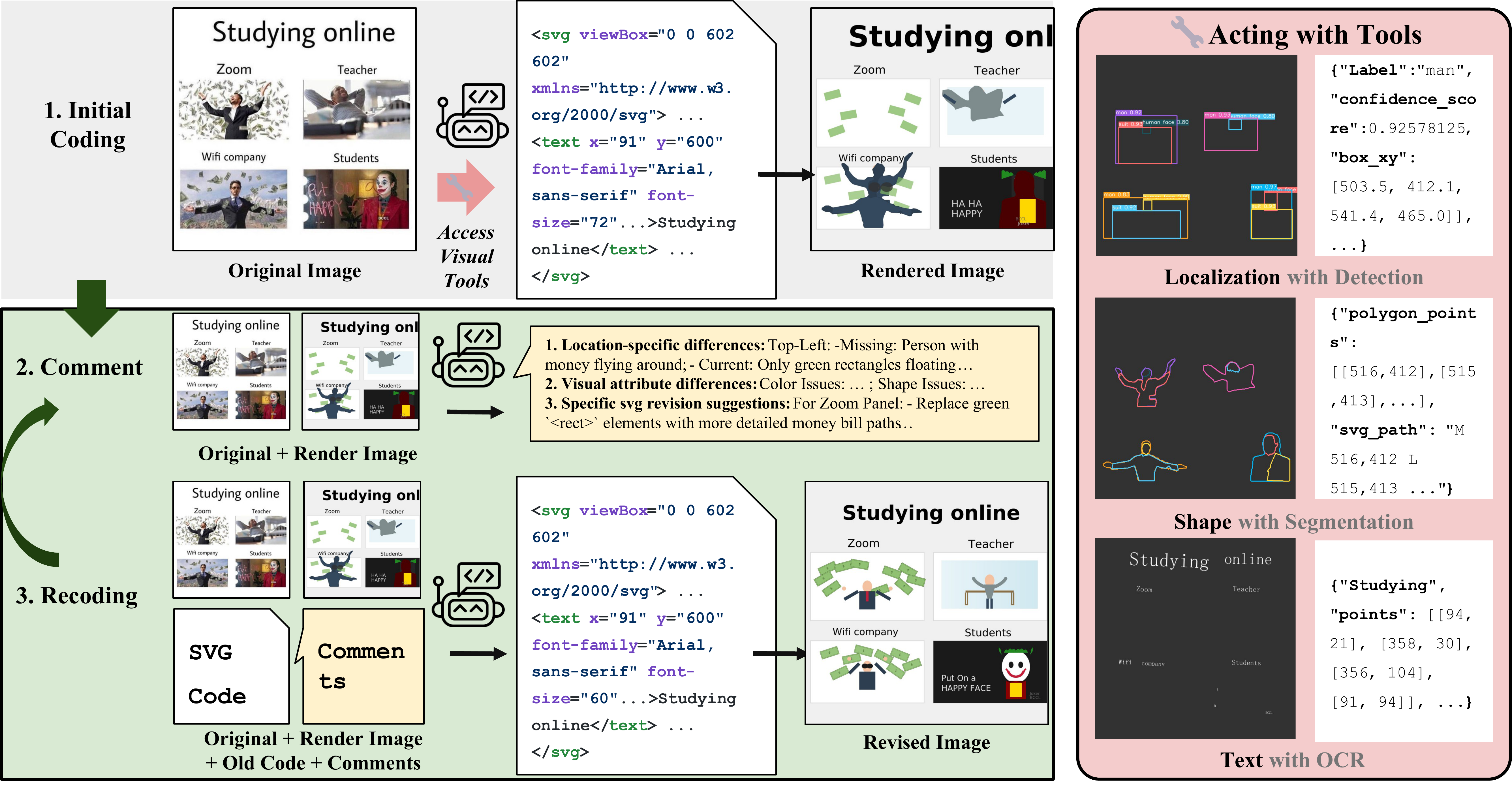}
\vspace{-1.5em}
\caption{\small{\textbf{Augmenting Coders with Test-time Revision \& Visual Tools.}  
\textbf{Left:} Thinking with Revision -- the model performs initial coding, comments on discrepancies between original and rendered images, and iteratively refines the SVG code.  
\textbf{Right:} Acting with Vision Tools -- external modules provide cues on categories, locations, shapes, colors, and text, which are translated into structured code signals to guide generation.  
These techniques yield more faithful and accurate renderings.}}
\vspace{-2em}
\label{fig:coder}
\end{figure}

\subsection{Act with Visual Tools}
Another limitation for Coder is capture the image fine-grained attribution such as boundary. Here we able the Coder to access additional visual tools to provide meta information to complement the generated SVG quality. We display part of tools with supple information in the right side of Fig.\ref{fig:coder}.

\textbf{Category.}  
Object categories are obtained from a detector~\cite{florence2} and provide the Coder with essential semantic labels. For example, a detected object can be annotated in SVG with an attribute like \texttt{id='bird'}. These labels serve as the basic prior for generation and are always combined with geometric cues like location or shape to describe each object more completely. 

\textbf{Location.} 
A key factor in reconstruction is capturing where objects appear in the image. To provide this information, we rely on bounding boxes predicted by Florence-2~\cite{florence2}, expressed as absolute coordinates $(x_1, y_1, x_2, y_2)$ together with the image width and height. These cues allow the Coder to anchor elements at the correct positions on the canvas, preserving the overall layout.

\textbf{Shape.}  
While regular geometric primitives are straightforward to express, a key challenge lies in representing irregular boundaries. To address this, we employ SAM-2~\cite{sam2} to generate segmentation masks that capture detailed object contours. These masks are then downsampled into sparse coordinate points through an adaptive simplification strategy, which reduces the number of vertices while keeping the overall area nearly unchanged. The resulting polygonal paths provide the Coder with compact yet faithful shape descriptions that complement category and location cues.

\textbf{Text.}  
Text often carries critical semantic information that cannot be replaced by shapes or colors.To incorporate this, We apply OpenOCR~\cite{openocr} to detect and transcribe text regions, and directly encode them into SVG using the native \texttt{<text>} tag, which preserves both content and visual attributes without the rendering issues of pixel-based methods. 
\section{Experiments}
\subsection{Baseline and Settings}
To comprehensively evaluate our proposed framework, we compare it against a wide range of proprietary and open-source models that represent the current state of the art in multimodal reasoning and code generation.
\textit{Proprietary models,} 
such as Claude-4.5-Sonnet, Claude-4-Opus and Claude-4-Sonnet, GPT-5, GPT-4.1, GPT-o3, GPT-4o, and GPT-4o-mini~\cite{GPT-4,GPT-4o}, as well as Gemini-2.5-Pro and Gemini-2.5-Flash~\cite{gemini}, and Seed-1.6-thinking~\cite{seed}. These models are widely recognized for their strong reasoning and multimodal capabilities, and thus provide competitive upper baselines for our benchmark.
\textit{Open-source models:} 
including LLaMA-4-Scout, Qwen3-VL, Qwen2.5-VL-72B and Qwen2.5-VL-7B~\cite{Qwen2.5-VL}, InternVL3.5-241B-A28B~\cite{Internvl3.5}, Intern-S1, InternVL3-78B~\cite{Internvl3}, MiniCPM-V-4.5~\cite{MiniCPM-V}, GLM-4.5V and GLM-4.1V-Thinking~\cite{GLM}, OmniSVG~\cite{omnisvg} and StarVector~\cite{starvector}. These baselines cover a diverse spectrum of model sizes and training paradigms, enabling a comparison between proprietary and open-source approaches.

\textbf{Evaluation settings.} 
Unless otherwise noted, all models are queried under a unified prompting interface with identical inputs to ensure fairness. The primary automatic evaluator is GPT-4o-mini, which provides consistent judgments across benchmarks. 

\subsection{Main Results}
\begin{table*}[]
\small
\centering
\label{main_table}
\resizebox{\textwidth}{!}{%
\begin{tabular}{l|rrr|rrrrrrr|r|rrr|r}
\toprule
\multirow{3}{*}{Model name} &
\multirow{3}{*}{\makecell{Success\\Rate (\%)}} &
\multirow{3}{*}{\makecell{SigLIP\\Score}} &
\multirow{3}{*}{\makecell{Code\\Token (\textit{K})}} &
\multicolumn{12}{c}{\metric} \\ 
 & & & & \multicolumn{7}{c|}{MM-Vet} & \multicolumn{1}{c|}{MMMU} & \multicolumn{3}{c|}{CV-Bench} & \multirow[c]{2}{*}{\makecell[c]{\textbf{Overall}}} \\
 & & & & Rec & Ocr & Know & Gen & Spat & Math & \textbf{Avg.} & \textbf{Avg.} & 2D & 3D & \textbf{Avg.} & \\
\midrule
\textcolor{gray}{Orig. Image (4o-mini)} & \textcolor{gray}{NA} & \textcolor{gray}{100.0} & \textcolor{gray}{NA} & \textcolor{gray}{60.5} & \textcolor{gray}{78.9} & \textcolor{gray}{58.5} & \textcolor{gray}{59.5} & \textcolor{gray}{70.9} & \textcolor{gray}{84.2} & \textcolor{gray}{67.1} & \textcolor{gray}{50.0} & \textcolor{gray}{77.4} & \textcolor{gray}{63.3} & \textcolor{gray}{70.3} & \textcolor{gray}{61.7} \\
\midrule
Claude-4.5-Sonnet & 99.1 & 66.8 & 1.9 & 29.7 & 57.6 & 11.9 & 17.0 & 57.3 & 52.7 & 39.0 & 42.5 & 50.4 & 55.0 & 52.7 & 43.1 \\
Claude-4-Opus & 98.2 & 65.9 & 1.5 & 30.4 & 52.3 & 13.9 & 18.5 & 49.5 & 50.4 & 37.5 & 42.5 & 41.6 & 58.3 & 49.9 & 41.7 \\
Claude-4-Sonnet & 98.2 & 65.5 & 1.6 & 31.8 & 51.2 & \underline{24.9} & \underline{27.9} & 44.8 & 34.6 & 37.8 & 39.0 & 49.0 & 53.3 & 51.2 & 41.1 \\
GPT-5 & 100.0 & \textbf{72.3} & 2.3 & \underline{33.9} & \textbf{64.9} & 20.5 & 23.8 & \textbf{60.5} & 65.4 & \underline{43.9} & 42.5 & 51.8 & \textbf{66.7} & \underline{59.2} & \underline{46.8} \\
GPT-4o & 100.0 & 60.6 & 0.6 & 23.1 & 58.4 & 12.7 & 17.0 & 51.3 & 60.4 & 35.0 & 44.5 & 29.3 & 50.0 & 39.6 & 39.0 \\
GPT-o3 & 100.0 & 64.1 & 1.1 & 31.3 & 55.2 & 17.7 & 19.7 & 48.5 & 61.5 & 39.8 & 39.0 & 47.4 & 56.7 & 52.1 & 42.2 \\
GPT-4.1 & 100.0 & 68.6 & 1.6 & 30.8 & 62.0 & 15.5 & 20.4 & 56.0 & 55.8 & 40.9 & 44.5 & 48.2 & \textbf{66.7} & 57.4 & 45.6 \\
GPT-4o-mini & 100.0 & 61.1 & \underline{0.4} & 20.7 & 58.4 & 13.2 & 18.9 & 46.8 & 63.5 & 33.5 & 44.5 & 27.7 & 48.3 & 38.0 & 37.9 \\
Gemini-2.5-Pro & 100.0 & 66.5 & 2.4 & 28.9 & 57.8 & 20.0 & 22.9 & 47.9 & \underline{68.5} & 39.1 & \underline{45.2} & \underline{56.1} & 56.7 & 56.4 & 44.7 \\
Gemini-2.5-Flash & 98.0 & 63.7 & 1.9 & 29.3 & 56.7 & 17.4 & 21.1 & 46.3 & 53.8 & 39.1 & 39.7 & 48.8 & 58.3 & 53.6 & 42.4 \\
Seed-1.6-Thinking & 100.0 & 62.8 & 1.6 & 18.9 & 46.5 & 8.1 & 11.9 & 44.1 & 38.5 & 28.7 & 43.2 & 45.3 & 51.7 & 48.5 & 37.5 \\
\midrule
Llama-4-Scout-17B-16E & 100.0 & 55.5 & 0.7 & 18.2 & 44.9 & 12.4 & 15.5 & 32.8 & 46.2 & 26.4 & 42.5 & 35.0 & 53.3 & 44.2 & 35.3 \\
Qwen3-VL-235B-A22B & 95.1 & 58.1 & 1.7 & 19.3 & 54.6 & 8.8 & 14.5 & 45.6 & 53.1 & 31.1 & 41.1 & 22.6 & 58.3 & 40.5 & 36.3 \\
Qwen2.5-VL-72B & 98.7 & 57.9 & \textbf{0.3} & 20.6 & 52.9 & 14.0 & 17.3 & 51.3 & 43.1 & 31.8 & 41.1 & 21.9 & 53.3 & 37.6 & 36.0 \\
Qwen2.5-VL-7B & 70.6 & 22.9 & 0.6 & 4.9 & 6.0 & 3.0 & 4.0 & 7.1 & 3.8 & 4.8 & 19.2 & 17.5 & 41.7 & 29.6 & 14.7 \\
InternVL3.5-241B-A28B & 100.0 & 60.2 & 1.0 & 20.4 & 52.4 & 11.9 & 15.7 & 39.2 & 42.3 & 31.1 & 43.8 & 45.3 & 50.0 & 47.6 & 38.7 \\
Intern-S1 & 100.0 & 60.0 & 1.0 & 24.7 & 56.8 & 12.1 & 16.0 & 51.2 & 41.9 & 35.2 & 41.1 & 46.8 & 55.0 & 50.9 & 40.4 \\
InternVL3-78B & 100.0 & 57.7 & 0.7 & 16.9 & 52.7 & 8.3 & 13.9 & 40.5 & 55.0 & 29.1 & 41.8 & 18.3 & 50.0 & 34.1 & 34.2 \\
MiniCPM-V-4.5 & 78.9 & 45.9 & 0.9 & 11.8 & 31.8 & 4.5 & 10.8 & 23.2 & 26.5 & 17.7 & 36.3 & 23.4 & 45.0 & 34.2 & 27.1 \\
GLM-4.5V & 99.8 & 63.8 & 1.6 & 22.4 & 54.4 & 7.1 & 15.6 & 46.0 & 56.9 & 33.1 & 40.4 & 43.1 & \textbf{66.7} & 54.9 & 40.1 \\
GLM-4.1V-Thinking & 100.0 & 61.7 & 1.2 & 21.1 & 52.0 & 10.4 & 13.7 & 44.8 & 58.8 & 31.9 & 43.2 & 37.9 & 56.7 & 47.3 & 38.8 \\
OmniSVG & 100.0 & 46.2 & 5.3 & 9.2 & 15.3 & 3.7 & 10.4 & 16.9 & 11.5 & 9.4 & 43.8 & 24.8 & 40.0 & 32.4 & 25.2 \\
StarVector & 8.3 & 18.1 & 1.3 & 0.0 & 3.4 & 0.0 & 1.6 & 4.4 & 0.0 & 1.5 & 6.8 & 0.0 & 0.0 & 0.0 & 2.8 \\
\midrule
\textbf{\coder~(Claude-4-Opus)} & 99.3	& \underline{71.0} & 2.0 & \textbf{46.6} & \underline{63.4} & \textbf{38.8} & \textbf{41.5} &	\underline{58.1} & \textbf{72.7} & \textbf{54.2$_{\text{+16.7}}$ } & \textbf{48.6$_{\text{+6.2}}$} & \textbf{57.7} & \underline{65.0} & \textbf{61.3$_{\text{+11.4}}$}	& \textbf{54.0$_{\text{+12.3}}$} \\
\bottomrule
\end{tabular}}
\caption{\small{\textbf{Main results on \our}~across various top-performing frontier VLM coders. Top half is the proprietary models, while the bottom half is the open-source model.
The best scores are in \textbf{bold} while the second best are in \underline{underline}.
The Overall score is calculated as an \textit{instance-weighted average} of the three subtasks (MM-Vet, MMMU, and CV-Bench) using their respective question counts.
}}
\label{table:main}
\vspace{-1.5em}
\end{table*}

In Tab.~\ref{table:main}, we evaluate full baselines on \our, reporting per-domain results—general, college, and vision-centric—and the overall average. We have the following observation.

\textbf{Stronger reasoning yield better visual coding scores.}
Closed-source models consistently outperform open-source counterparts across benchmarks. GPT-5 sets the strongest baseline with the top SigLIP score (72.3) and the highest CodeVQA overall (46.8), showing robust performance on both similarity and reasoning metrics. This pattern indicates that stronger reasoning ability translates into better VCode performance—\ie~models that reason well produce more faithful symbolic renderings. We also observe a positive correlation between semantic similarity (SigLIP) and CodeVQA.

\textbf{Challenges across different dimensions.}
\textit{(i) Best performer still trails the original-image upper bound.}
Even the best SVG result—GPT-5 at 46.8—remains well below the raw-image upper bound (61.7), indicating substantial headroom. This confirms that the task is sufficiently challenging and that symbolic representation still has ample room for improvement.
\textit{(ii) SVG specialist underperforms.} OmniSVG and StarVector-8B ranks last due to the low success rate for long context outputs, highlighting \our’s difficulty and the gap between neatly authored SVG corpora and SVGs derived from natural images. 
\textit{(iii) Knowledge is hardest.} In MM-Vet, the Knowledge dimension is consistently the lowest, reflecting the compounded challenge of recalling facts and then encoding them faithfully in SVG (\eg~historical entities). 
\textit{(iv) Professional disciplines are hard to differentiate.} On MMMU, models cluster within a narrow, modest band, and most fail the more demanding disciplinary settings.
\textit{(v) Vision-centric perception is tough.} CV-Bench scores hover near the low (randomly by 50\%), especially on 3D relations (depth or spatial). Even with \coder, improvements are meaningful but leave substantial headroom.

\textbf{Absolute gains with \coder.} Built on Claude-4-Opus, \coder~lifts Overall from 41.7 to 54.0 (+12.3) via revision and vision-tool assistance, improving all three domains—demonstrating an effective enhancement for visual-centric coding.

\textbf{Code token length is highly correlated with expressiveness.} Models that emit short SVGs underperform (\eg~0.3K by Qwen-2.5-VL). By contrast, stronger models (GPT-5, Gemini-2.5-Pro) produce substantially longer sequences (often \(>\!2\)K tokens) and attain higher scores. Length is not sufficient on its own, but performance scales with usable context, highlighting long-context reasoning and generation as a central bottleneck for visual-centric coding.

\subsection{Key Ablations}
\begin{figure}[htbp]
\centering
\begin{minipage}{0.64\linewidth} 
\centering
\renewcommand{\arraystretch}{1}
\large
\setlength{\tabcolsep}{3pt} 
\scriptsize
\begin{tabular}{l|c|ccccccc}
\toprule
\multirow{2}{*}{Variant} & \multirow{2}{*}{\makecell{SigLIP\\Score}} & \multicolumn{7}{c}{CodeVQA--MMVet} \\ 
 & & Rec & Ocr & Know & Gen & Spat & Math & \textbf{Avg.} \\
\midrule
Claude-4-Opus         &  65.6  & 30.4 & 52.3 & 13.9 & 18.5 & 49.5 & 50.4 & 37.5 \\
$+$Loc. \& {C.}          & \underline{70.8} & 29.7 & 60.3 & 17.5 & 22.9 & 54.9 & 46.2 & 39.7 \\
$+$Loc. \& {C.} \& {S.}     & 71.5  & 33.4 & 62.7 & 19.3 & 25.1 & \underline{63.1} & 64.2 & 43.3 \\
$+$Text            & 69.9 & 30.4 & 59.5 & 19.2 & 21.5 & 56.8 & \underline{65.4} & 41.5 \\
\textbf{$+$Full vision tools}             & \textbf{71.6} &\textbf{46.0} & \underline{64.4} & \textbf{40.8} & \textbf{43.0} & 61.6 & \textbf{72.7} & \textbf{54.1} \\
\bottomrule
\end{tabular}
\captionof{table}{\small{\textbf{Effects by vision tools modules,} where Loc. denotes Location, {C.} denotes Category, and {S.} denotes Shape.}}
\vspace{-1em}

\label{ablation:vcoder}
\end{minipage}
\hfill
\begin{minipage}{0.35\linewidth}
\centering
\includegraphics[width=\linewidth]{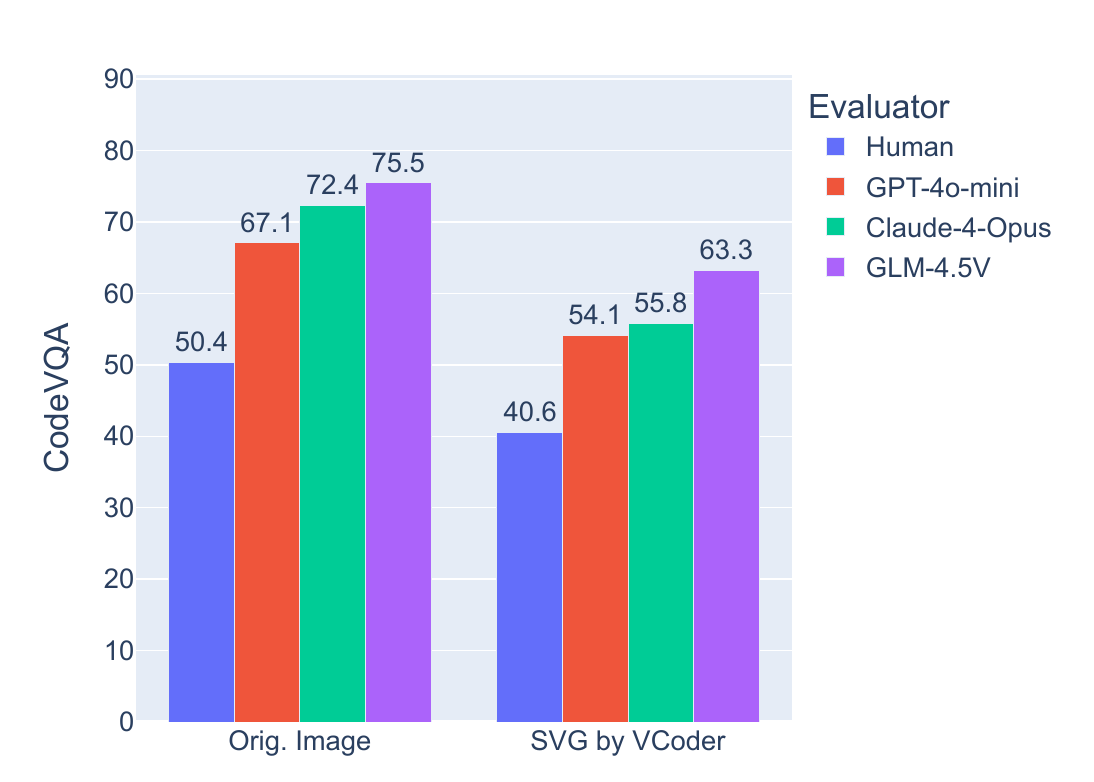}
\vspace{-2em}
\captionof{figure}{\small{\textbf{Effects by different policy during evaluation}}}
\vspace{-1em}
\label{ablation:evaluator}
\end{minipage}
\end{figure}

\textbf{Effects of Vision tools.}
Ablations in Tab.~\ref{ablation:vcoder} reveal three takeaways:
\textit{(i)} Adding fine-grained cues (location, category, shape) yields steady gains; shape is especially helpful for spatial reasoning (Spat.), even without large changes in SigLIP, indicating structural benefits.
\textit{(ii)} Text cues help, with the full visual-tool ensemble provides the largest overall improvement.
\textit{(iii)} Together, all vision tools yield a 16.6-point improvement over Claude-4-Opus, implying the strong potential of the Coder itself to autonomously call tools and leverage contextual information for code generation.

\textbf{Effects across Policies and Human studies.}
Fig.~\ref{ablation:evaluator} shows the performance differences across policy $\phi$, including humans. On the original images, all models substantially surpass human perception and reasoning (50.4 for humans vs.~75.5 for GLM-4.5V). When evaluated on SVG representations, all models exhibit a noticeable performance drop, even with the human score decreasing to 40.6. 
Interestingly, both humans and VLMs exhibit a form of alignment when interpreting symbolic representations. Although abstraction inevitably leads to information loss compared to original visual inputs, VLMs demonstrate a comparable ability to reason from such high-level representations, suggesting that their potential in this aspect is on par with human understanding.

\begin{figure}[htbp]
\centering
\begin{minipage}{0.58\linewidth}
\centering
\renewcommand{\arraystretch}{1.6} 
\captionsetup{justification=centering}
\setlength{\tabcolsep}{1.5pt}       
\scriptsize
\begin{tabular}{l|c|c|cccc}
\toprule
{Variant} & {SigLIP} & 
{Code} &
\multicolumn{4}{c}{CodeVQA} \\
 & Score & Token  & {MM-Vet} & {MMMU} & {CV-Bench} & \textbf{Overall} \\
\midrule
Img2SVG &  65.6  & 1.5K & 37.5 & \textbf{43.2} & 52.3 & 42.5 \\
Img2SVG-Thinking & \textbf{69.8} & 1.6K & \underline{38.2} & \underline{42.5} & \textbf{53.7} & \underline{43.5} \\
Img2Text2SVG &  \underline{68.5}  & 1.8K & \textbf{43.0} & \textbf{43.2} & \underline{55.6} & \textbf{46.4} \\
\bottomrule
\end{tabular}
\captionof{table}{\small{\textbf{Effects by different input modes} of Claude-4-Opus}}
\label{ablation:input}
\end{minipage}
\hfill
\begin{minipage}{0.36\linewidth}
\centering
\includegraphics[width=\linewidth]{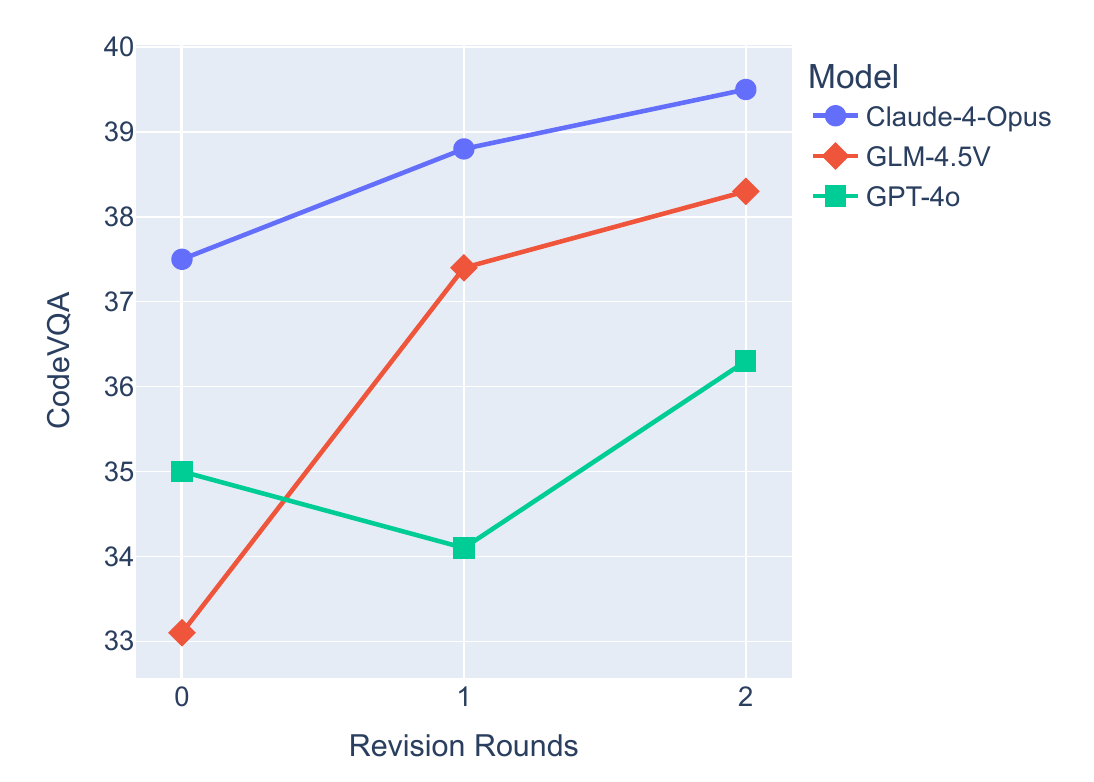}
\vspace{-2em}
\captionof{figure}{\small{\textbf{Effect of revision round.}}}
\vspace{-1em}
\label{ablation:revision}
\end{minipage}
\end{figure}

\textbf{Effects by Revision.}
In Fig.~\ref{ablation:revision}, we examine the impact of our revision strategy. Both Claude and GLM-4.5V benefit from the first revision, with GLM-4.5V showing the most substantial gains---likely due to its built-in ``thinking mode,'' which excels at difference analysis and refinement. In contrast, GPT-4o initially struggles during the first revision but continues to improve in later rounds, implying that effective revision relies on a strong reasoning foundation.

\begin{figure*}[!h]
  \centering
  \begin{subfigure}[b]{0.24\textwidth}
    \includegraphics[width=\linewidth]{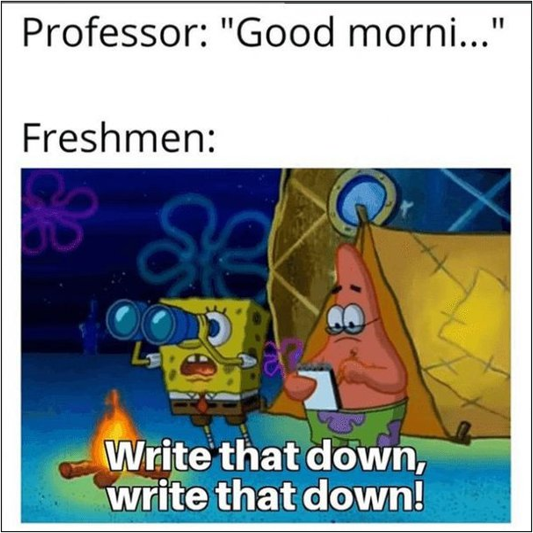}
    \caption{Original image}
    \label{fig:ex1}
  \end{subfigure}\hfill
  \begin{subfigure}[b]{0.24\textwidth}
    \includegraphics[width=\linewidth]{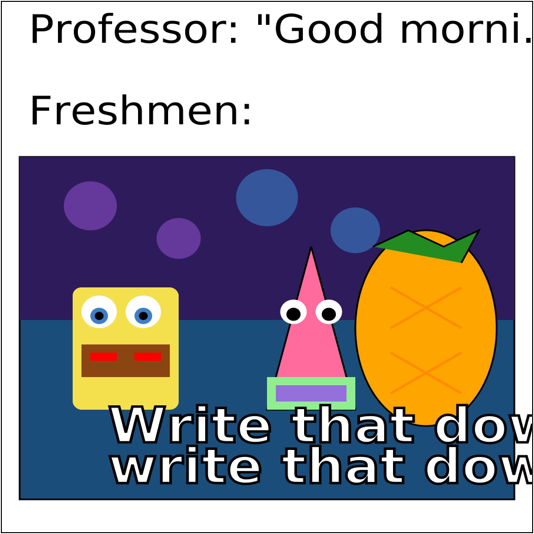}
    \caption{Initial rendered}
    \label{fig:ex2}
  \end{subfigure}\hfill
  \begin{subfigure}[b]{0.24\textwidth}
    \includegraphics[width=\linewidth]{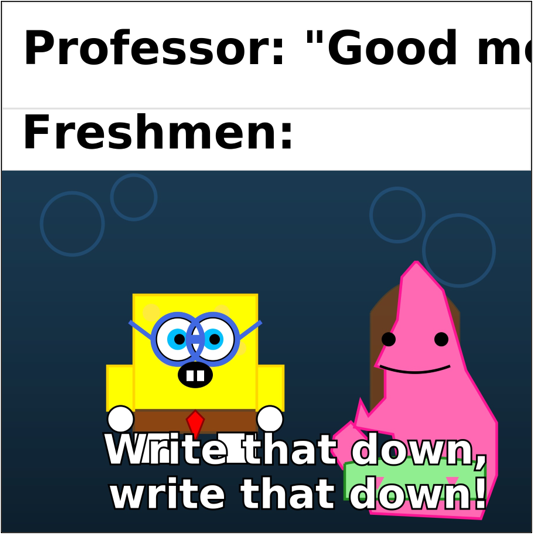}
    \caption{w. visual tools}
    \label{fig:ex3}
  \end{subfigure}\hfill
  \begin{subfigure}[b]{0.24\textwidth}
    \includegraphics[width=\linewidth]{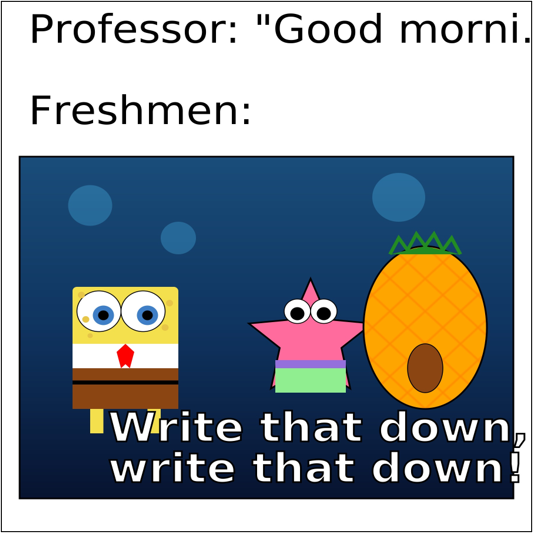}
    \caption{w. revision}
    \label{fig:ex4}
  \end{subfigure}

  \vspace{1mm}

  \begin{subfigure}[b]{0.24\textwidth}
    \includegraphics[width=\linewidth]{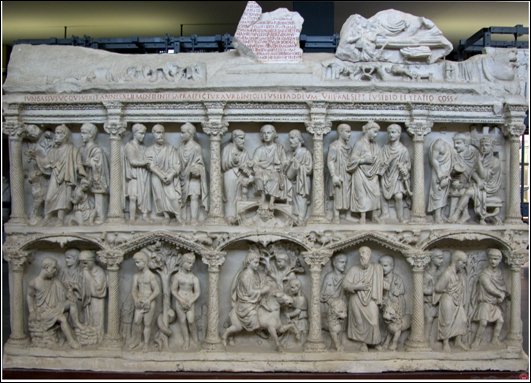}
    \caption*{MMMU's example}
    \label{fig:ex5}
  \end{subfigure}\hfill
  \begin{subfigure}[b]{0.24\textwidth}
    \includegraphics[width=\linewidth]{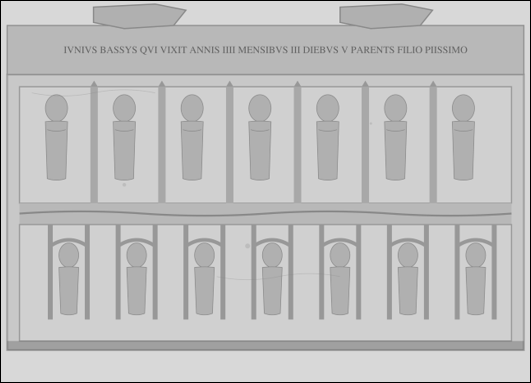}
    \caption*{Rendered by \coder}
    \label{fig:ex6}
  \end{subfigure}\hfill
  \begin{subfigure}[b]{0.24\textwidth}
    \includegraphics[width=\linewidth]{figures/ex7.png}
    \caption*{CV-Bench's example}
    \label{fig:ex7}
  \end{subfigure}\hfill
  \begin{subfigure}[b]{0.24\textwidth}
    \includegraphics[width=\linewidth]{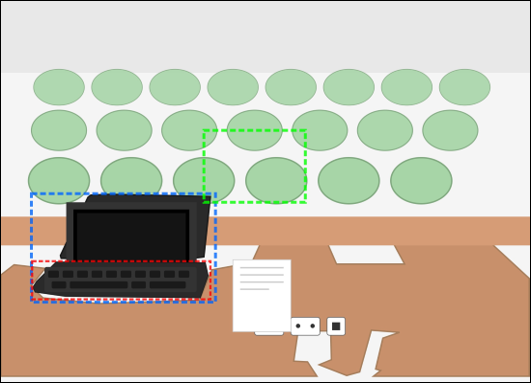}
    \caption*{Rendered by \coder}
    \label{fig:ex8}
  \end{subfigure}

  \caption{\small{\textbf{Qualitative examples from \our.} \textbf{Top row (a–d):} an internet meme rendered progressively by initial decoding, visual-tool assistance, and revision. \textbf{Bottom row:} challenge samples from MMMU (Art-Theory) and CV-Bench (3D), alongside their SVG renderings by \coder.}}
  \vspace{-1.5em}
  \label{fig:example}
\end{figure*}

\textbf{Effects by Visual v.s. Textual query.}
In Tab.~\ref{ablation:input}, we examine the impact of input modality. Using raw images (\ie~{Img2SVG}) gives the weakest results, suggesting that current coders are poorly adapted to direct visual input. 
Notably, even with deep thinking enabled (\ie~{Img2SVG-Thinking}), performance remains low, underscoring the difficulty of visual-centric coding and the gap between language-driven and vision-driven code generation.
By contrast, translating images into linguistic captions before coding (\ie~{Img2Text2SVG}) achieves the best performance, highlighting the benefit of language as an intermediate representation. 

\subsection{Qualitative Analysis}
Fig.~\ref{fig:example} presents qualitative results by comparing origina image and the rendered image by \coder.
\textbf{Top row.} Across the four stages, the initial decoding misses layout and semantics. Adding \emph{visual tools} recovers key geometry (\eg~the starfish character’s triangular body and facial features), while \emph{revision} corrects fine details (character proportions, text alignment, spacing), yielding a rendering that closely matches the meme’s structure and intent. \textbf{Bottom row.} \coder~produces SVGs that are both more faithful to the source and more interpretable for downstream reasoning. The left example (MMMU) is knowledge-intensive: accurately depicting a multi-panel historical relief requires domain cues and fine structural abstraction, where base models often collapse detail. The right example (CV-Bench) is vision-centric: success hinges on \emph{visually grounded prompts} that localize and size objects correctly (\eg~monitor in front of keyboard, receding rows of chairs), after which revision tightens residual misalignments. These examples underscore the challenges by \our.
\section{Conclusion}
We introduced \our, offering a new perspective on multimodal coding by benchmarking multimodal understanding with SVG as a visual representation, along with \metric\ to assess symbolic fidelity through QA over rendered SVGs. Our study shows that frontier VLMs struggle to produce faithful SVGs despite strong linguistic reasoning, revealing a persistent gap between language- and vision-centric coding. To address this, we proposed \coder, which integrates {Test-time Revision} and {Acting with Visual Tools}, yielding substantial improvements. Human studies underscore the potential of symbolic visual coding as a pathway toward more human-aligned multimodal intelligence. Future work can focus on developing end-to-end vision–language coders with scalable training data to enable more faithful symbolic representations.

\bibliography{main}
\bibliographystyle{unsrt}

\clearpage
\appendix              
\startcontents[app]    
\printcontents[app]{l}{1}

\section{Implement Details}
We implement our model using the PyTorch framework on an NVIDIA RTX 4090 GPU with 24GB of memory. The maximum output length is set to 16,384 tokens, while for the Qwen2.5-VL models we use 8,192 tokens.

For evaluation, different protocols are used depending on the benchmark. 
In MM-Vet, we employ \texttt{gpt-4-0613} as the evaluator to score model responses. 
In CV-Bench and MMMU experiments, we adopt a rule-based string matching parser to determine correctness.

For SigLip2, we use the \texttt{siglip2-so400m-patch14-384}.
The token cost reported in our tables is measured using the \texttt{tiktoken} library with the \texttt{cl100k\_base} encoding. 

It is worth noting that in the \emph{img2svg} experiments, StarVector cannot take textual prompts as input. 
It directly performs image-to-SVG generation. 

\begin{algorithm}[!h]
\caption{Test-time Revision}
\label{alg:iterative-revision}
\begin{algorithmic}[1]
\small
\State \textbf{Input:} Coder $\psi$, an image $\mathcal{V}$, initial rendering $\widetilde{\mathcal{V}}^{(0)}$, initial SVG code $\mathcal{C}^{(0)}$, iteration number $T$
\State \textbf{Output:} Refined rendering $\widetilde{\mathcal{V}}^{(T)}$
\For{$t = 0 \to \left(T-1\right)$}
    \State Comment the difference: $\Delta^{(t)} \gets \psi\left(\mathcal{V}, \widetilde{\mathcal{V}}^{(t)}\right)$
    \State Generate revised SVG code: $\mathcal{C}^{(t+1)} \gets \psi(\mathcal{V}, \widetilde{\mathcal{V}}^{(t)}, \Delta^{(t)}, \mathcal{C}^{(t)})$
    \State Update reconstruction: 
    $\widetilde{\mathcal{V}}^{(t+1)} \gets \mathrm{Render}(\mathcal{C}^{(t+1)})$
\EndFor
\State \Return $\widetilde{\mathcal{V}}^{(T)}$
\end{algorithmic}
\end{algorithm}

\section{Experiments Ablations}
\subsection{Effects by SigLip v.s. DINO}
\begin{table*}[htbp]
\centering
\setlength{\tabcolsep}{3pt} %
\resizebox{\textwidth}{!}{%
\begin{tabular}{lcccccccccccc}
\toprule
Metric & Claude-4.5-Sonnet & Claude-4-Opus & Claude-4-Sonnet & GPT-5 & GPT-4o & GPT-o3 & GPT-4.1 & GPT-4o-mini & Gemini-2.5-Pro & Gemini-2.5-Flash & Seed-1.6-Thinking & VCoder \\
\midrule
SigLip2 & 66.9 & \underline{67.2} & 66.9 & \textbf{70.1} & 60.1 & 64.9 & 66.9 & 58.6 & 66.4 & 64.3 & 62.6 & 72.3 \\
DINO-V2 & 26.2 & 26.1 & 24.2 & \textbf{30.4} & 16.5 & 22.4 & 26.0 & 14.3 & \underline{27.2} & 22.5 & 20.2 & 33.0 \\
\bottomrule
\end{tabular}}
\caption{\small{\textbf{Effect by different feature extractor.} For each metric, the best results are highlighted in \textbf{bold}, and the second-best results are \underline{underlined}. As shown, the DINO reach lower score compare with SigLip2, as it more focus on low-level visual representation. While SigLip2 focus on more on semantic space.}}
\end{table*}

\subsection{Effect by revision round on MM-Vet}

\begin{table*}[htbp]
\centering
\label{tools_ablation}
\footnotesize
\begin{tabular}{l|c|ccccccc}
\toprule
Models & Round & Rec & Ocr & Know & Gen & Spat & Math & Avg. \\
\midrule
\multirow{3}{*}{Claude-4-Opus} 
 & 0 & \textbf{30.4} & 52.3 & 13.9 & \underline{18.5} & 49.5 & 50.4 & 37.5 \\
 & 1 & 29.0 & \textbf{54.3} & \textbf{18.9} & \textbf{21.7} & \textbf{56.0} & \underline{53.1} & \underline{38.8} \\
 & 2 & \underline{29.6} & \underline{54.0} & \underline{16.9} & 14.5 & \underline{53.1} & \textbf{55.4} & \textbf{39.5} \\
\midrule
\multirow{3}{*}{GLM-4.5V} 
 & 0 & 22.4 & 54.4 & 7.1  & \underline{15.6} & 46.0 & \textbf{56.9} & 33.1 \\
 & 1 & \textbf{26.5} & 58.3 & \textbf{14.5} & \textbf{20.0} & \textbf{54.4} & 50.0 & \underline{37.4} \\
 & 2 & \underline{24.5} & \textbf{65.7} & \underline{10.4} & \underline{15.6} & \underline{53.3} & \underline{55.8} & \textbf{38.3} \\  
\midrule
\multirow{3}{*}{GPT-4o} 
 & 0 & 23.1 & \underline{58.4} & 12.7 & \underline{17.0} & \textbf{51.3} & \underline{60.4} & \underline{35.0} \\
 & 1 & \underline{23.7} & 53.5 & 12.0 & 15.7 & 46.5 & 53.5 & 34.1 \\
 & 2 & \textbf{25.0} & \textbf{60.0} & \textbf{14.2} & \textbf{18.9} & \underline{50.7} & \textbf{56.9} & \textbf{36.3} \\
\bottomrule
\end{tabular}
\caption{\small{\textbf{Effect by revision (round) on MM-Vet.} For each model, the best results are highlighted in \textbf{bold}, and the second-best results are \underline{underlined}.}}
\label{ablation:revision_round_table}
\end{table*}

\subsection{Effects by different policy during evaluation on MM-Vet}

\begin{table*}[h]\centering
\small
\label{tab:evaluator_results}
\begin{tabular}{l|l|ccccccc}
\toprule
Setting & Evaluator & Rec & Ocr & Know & Gen & Spat & Math & Avg. \\
\midrule
\multirow{4}{*}{Ori} 
 & GPT-4o-mini   & 60.5 & 78.9 & \underline{58.5} & \underline{59.5} & 70.9 & \underline{84.2} & \underline{67.1} \\
 & Human     & 40.8	& 67.6 & 20.4 & 21.5 & 69.5 & 74.8 & 50.4 \\
 & Claude-4-Opus & \textbf{68.1} & \underline{79.3} & \textbf{59.0} & 57.9 & \textbf{82.1} & 72.7 & 72.4 \\
 & GLM-4.5V      & \underline{67.4} & \textbf{87.1} & 56.5 & \textbf{60.0} & \underline{80.0} & \textbf{96.2} & \textbf{75.5} \\
\midrule
\multirow{4}{*}{VCoder} 
 & GPT-4o-mini   & 46.0 & 64.4 & \underline{40.8} & \underline{43.0} & 61.6 & 72.7 & 54.1 \\
 & Human         & 30.5 &	55.8	&14.4	&15.6	&59.8	&61.4	& 40.6 \\
 & Claude-4-Opus & \underline{43.7} & \textbf{76.3} & 37.0 & 41.2 & \underline{68.4} & \underline{76.5} & \underline{55.8} \\
 & GLM-4.5V      & \textbf{54.3} & \underline{73.6} & \textbf{48.2} & \textbf{49.0} & \textbf{73.6} & \textbf{84.2} & \textbf{63.3} \\
\bottomrule
\end{tabular}
\caption{\small{\textbf{Evaluation results of different evaluators on Ori vs VCoder.} 
For each setting (Ori and VCoder), the best results are highlighted in \textbf{bold}, and the second-best results are \underline{underlined}.}}
\end{table*}

\section{Prompt Template}

\subsection{Img2SVG}
\begin{prompt}{Img2SVG}
Convert this image to SVG code following these rules: \\[3pt]
\textbf{CRITICAL REQUIREMENTS:} \\[3pt]
– Output only pure SVG code, no markdown blocks or explanations. \\
– Start with \verb|<svg viewBox="..." xmlns="http://www.w3.org/2000/svg">| and end with \verb|</svg>|. \\
– Use only native SVG elements (no external images or links). \\
– Include \texttt{viewBox} to ensure all elements are visible and auto-scale properly. \\
– Calculate appropriate \texttt{viewBox} dimensions to contain all content with some padding. \\[3pt]
\textbf{Generate the SVG now.}
\end{prompt}

\subsection{Img2Text2SVG}

\begin{prompt}{Img2Text2SVG – Stage 1: Image Captioning}
Please provide a detailed and accurate description of this image. Focus on: \\[3pt]
1. Main objects, shapes, and elements \\
2. Colors, textures, and visual properties \\
3. Spatial relationships and positioning \\
4. Style and artistic characteristics \\
5. Any text, symbols, or specific details \\[3pt]
Be precise and comprehensive. This description will be used to recreate the image as an SVG. 
Include geometric details, proportions, and layout information that would be necessary for accurate reproduction.
\end{prompt}

\begin{prompt}{Img2Text2SVG – Stage 2: SVG Generation}
Based on the following description, generate clean and accurate SVG code: \\[3pt]
\texttt{\{description\}} \\[3pt]
\textbf{CRITICAL REQUIREMENTS:} \\[3pt]
1. Output ONLY complete SVG code, no explanations or other text. \\
2. Use appropriate dimensions (e.g., \verb|viewBox="0 0 400 400"| or similar). \\
3. Include all elements described with accurate colors, shapes, and positioning. \\
4. Use clean, well-structured SVG syntax. \\
5. Ensure the SVG is self-contained and complete. \\
6. Start with \verb|<svg viewBox="..." xmlns="http://www.w3.org/2000/svg">| and end with \verb|</svg>|. \\
7. Use precise geometric shapes and paths where appropriate. \\
8. Match colors and proportions as closely as possible to the description. \\[3pt]
\textbf{Generate the SVG now.}
\end{prompt}

\subsection{Img2SVG-Thinking}
\begin{prompt}{Img2SVG – Thinking}
Let's analyze this image and create an SVG representation through a structured thinking process. \\[3pt]
\textbf{Step-by-step analysis:} \\
1. Visual Decomposition \\
– What are the main visual elements? \\
– What geometric shapes can be identified? \\
– What are the key colors and their relationships? \\[3pt]
2. Structural Analysis \\
– How are elements arranged and layered? \\
– What are the proportions and spatial relationships? \\
– Are there any repeating patterns or symmetry? \\[3pt]
3. SVG Implementation Strategy \\
– Which SVG elements best represent each component? \\
– What's the optimal drawing order? \\
– How to handle complex shapes and gradients? \\[3pt]
4. Technical Considerations \\
– What viewport dimensions are appropriate? \\
– How to ensure scalability and responsiveness? \\
– What optimizations can be applied? \\
\textbf{After your analysis, provide:} \\
1. Your complete reasoning process \\
2. The final SVG code implementation \\
\textbf{Requirements for SVG output:} \\
– Must be complete and self-contained. \\
– Include all necessary attributes and elements. \\
– Start with \verb|<svg| tag and end with \verb|</svg>|. \\
– Use appropriate \texttt{viewBox} and dimensions. \\
\textbf{Please proceed with the analysis and generation.}
\end{prompt}

\subsection{Visual Tools}
\begin{prompt}{Visual Tools – System Prompt}
You are a helpful assistant that converts images into clean, complete SVG vector graphics. \\[3pt]
Your primary task is img2svg conversion for Visual Question Answering. You have access to two types of metadata to assist with precision: \\[3pt]
\textbf{METADATA AVAILABLE:} \\
– OCR metadata: Text regions with precise 4-point quadrilaterals for accurate text placement. \\
– Object detection metadata: Object boundaries with labels, confidence scores, and \texttt{svg\_path} outlines. \\[3pt]
\textbf{SPECIAL CASE HANDLING (Hint Strategy):} \\
Sometimes, an image may depict a person, character, or artwork where fine details like facial features or texture could be lost during vectorization. Examples include: \\
– A recognizable public figure such as a scientist or political leader \\
– A well-known fictional character from popular culture \\
– A famous painting or portrait by a specific artist \\[3pt]
If the subject in the image is of this nature and important identity cues might be lost: \\
– Preserve recognizability by including visual hints such as characteristic clothing, accessories, environment, or symbolic elements. \\
– When confident, you may add a \verb|<text>| element near the subject that provides their commonly known name, the name of the associated work or series, or the title/creator of an artwork. \\[3pt]
If the subject does not fit these examples or is not clearly recognizable: \\
– Generate a clean SVG with no extra text labels. \\
– Focus on accurate shapes, proportions, and composition. \\[3pt]
\textbf{METADATA INTEGRATION:} \\
1) Text rendering: Use OCR quadrilaterals as authoritative coordinates for text placement. Render literal text strings with appropriate transforms for rotation/skew. \\
2) Object boundaries: Use detection \texttt{svg\_paths} as authoritative contours. Infer fill/stroke colors and add internal details within these boundaries. \\
3) Background reconstruction: Fill in unlabeled regions using native SVG primitives. \\[3pt]
\textbf{PROCESSING PRIORITY:} \\
1. Use provided metadata for precise positioning (OCR quads, detection paths). \\
2. Apply hint strategy for recognizable subjects. \\
3. Reconstruct missing background/unlabeled areas. \\
4. Ensure proper layering and visual completeness. \\[3pt]
\textbf{OUTPUT REQUIREMENTS:} \\
– Output only pure SVG code, no markdown blocks or explanations. \\
– Start with \verb|<svg viewBox="..." xmlns="http://www.w3.org/2000/svg">| and end with \verb|</svg>|. \\
– Use only native SVG elements (no external images or links). \\
– Include \texttt{viewBox} to ensure all elements are visible and auto-scale properly. \\
– Do not include explanations or commentary. \\[3pt]
This SVG will be used in a Visual Question Answering task, so ensure the output retains as much semantic identity as possible when visual details are reduced.
\end{prompt}

\begin{prompt}{Visual Tools – User Prompt}
Image dimensions: \{W\}x\{H\} \\

\textbf{METADATA:} \\
\texttt{\{metadata\_json\}} \\[6pt]

Generate the complete SVG with precise metadata integration and appropriate hint strategy for recognizable subjects.
\end{prompt}

\subsection{Revision}
\begin{prompt}{Revision – Stage 1: Visual Difference Analysis}
Compare the original image (first) with the SVG-rendered image (second) and identify \textbf{specific differences} for SVG code revision. \\[3pt]
\textbf{Focus on identifying:} \\[3pt]
1. LOCATION-SPECIFIC DIFFERENCES: \\
– Which areas or regions differ (top-left, center, bottom-right, etc.). \\
– Missing or extra elements in specific positions. \\[3pt]
2. VISUAL ATTRIBUTE DIFFERENCES: \\
– Color mismatches (specify which elements and what colors). \\
– Shape distortions (which shapes are wrong and how). \\
– Size or proportion issues (which elements are too big or too small). \\
– Position or alignment problems. \\[3pt]
3. SPECIFIC SVG REVISION SUGGESTIONS: \\
– Which SVG elements need modification (\texttt{circle}, \texttt{path}, \texttt{rect}, etc.). \\
– What attributes to change (\texttt{fill}, \texttt{stroke}, \texttt{cx}, \texttt{cy}, \texttt{width}, \texttt{height}, \texttt{d}, etc.). \\
– Specific color values or coordinate adjustments needed. \\[3pt]
Format your response as actionable SVG revision instructions.
\end{prompt}

\begin{prompt}{Revision – Stage 2: SVG Code Correction}
You are an SVG code specialist. Based on the visual analysis and comparison between the original image and the current SVG rendering, make \textbf{specific code modifications} to fix identified issues. \\[3pt]
\textbf{VISUAL ANALYSIS FINDINGS:} \\
\texttt{\{optimization\_goals\}} \\[3pt]
\textbf{CURRENT SVG CODE:} \\
\texttt{\{current\_svg\_code\}} \\[3pt]
\textbf{INSTRUCTIONS:} \\
1. Analyze the current SVG code structure and elements. \\
2. Based on the visual analysis findings, identify which specific SVG elements need modification. \\
3. Make precise changes to fix the identified issues: \\
   – Adjust colors (\texttt{fill}, \texttt{stroke} attributes). \\
   – Fix shapes and paths (modify \texttt{d} attributes, coordinates). \\
   – Correct sizes and positions (\texttt{width}, \texttt{height}, \texttt{cx}, \texttt{cy}, \texttt{x}, \texttt{y}). \\
   – Add missing elements or remove incorrect ones. \\
4. Output \textbf{only} the complete revised SVG code. \\
5. Ensure all modifications directly address the issues mentioned in the analysis. \\
6. Start with \verb|<svg| and end with \verb|</svg>|. \\[3pt]
\textbf{Revised SVG code:}
\end{prompt}

\section{More Visualizations}
\subsection{VCoder vs. Baselines}
In this section, we present qualitative comparisons between VCoder and baseline models on representative examples from three benchmarks. For each case, we display: (a) the original reference image, (b) the output generated by VCoder, and (c–d) the visual results produced by the two strongest baseline models. These comparisons clearly illustrate VCoder's superior ability to faithfully interpret and reconstruct visual content while preserving semantic consistency with the reference images.
\subsubsection{MM-VET}
\noindent
\begin{center}
  \begin{minipage}[b]{0.24\textwidth}\centering\includegraphics[width=\linewidth]{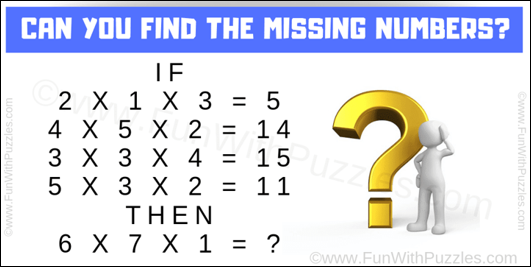}\\(a) Original image\end{minipage}\hfill
  \begin{minipage}[b]{0.24\textwidth}\centering\includegraphics[width=\linewidth]{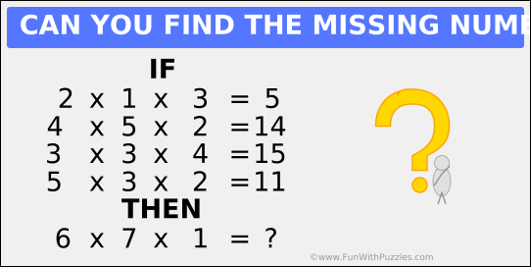}\\(b) VCoder\end{minipage}\hfill
  \begin{minipage}[b]{0.24\textwidth}\centering\includegraphics[width=\linewidth]{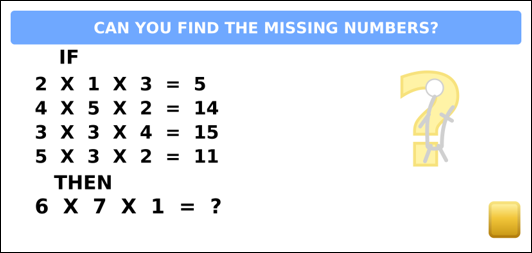}\\(c) GPT-5\end{minipage}\hfill
  \begin{minipage}[b]{0.24\textwidth}\centering\includegraphics[width=\linewidth]{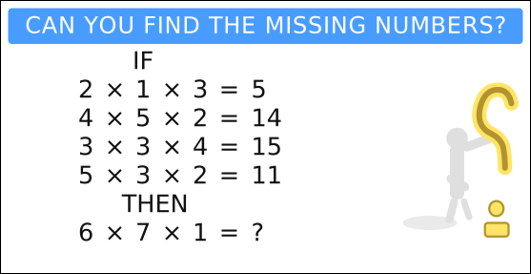}\\(d) GPT-4.1\end{minipage}
\end{center}
\begin{center}
  \textbf{Question:} Find the pattern of how the "X" operator is redefined, and answer the given equation in the image. \textbf{Answer:} 13
\end{center}

\noindent
\begin{center}
  \begin{minipage}[b]{0.24\textwidth}\centering\includegraphics[width=\linewidth]{figures/mmvet_part2/21.png}\\(a) Original image\end{minipage}\hfill
  \begin{minipage}[b]{0.24\textwidth}\centering\includegraphics[width=\linewidth]{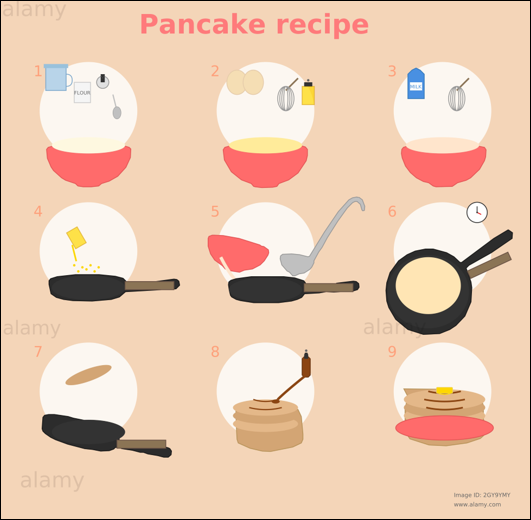}\\(b) VCoder\end{minipage}\hfill
  \begin{minipage}[b]{0.24\textwidth}\centering\includegraphics[width=\linewidth]{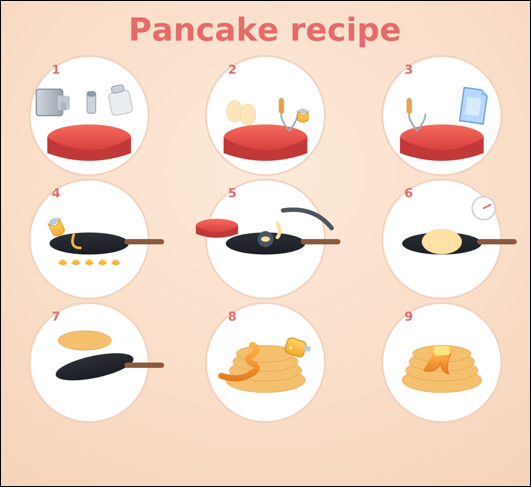}\\(c) GPT-5\end{minipage}\hfill
  \begin{minipage}[b]{0.24\textwidth}\centering\includegraphics[width=\linewidth]{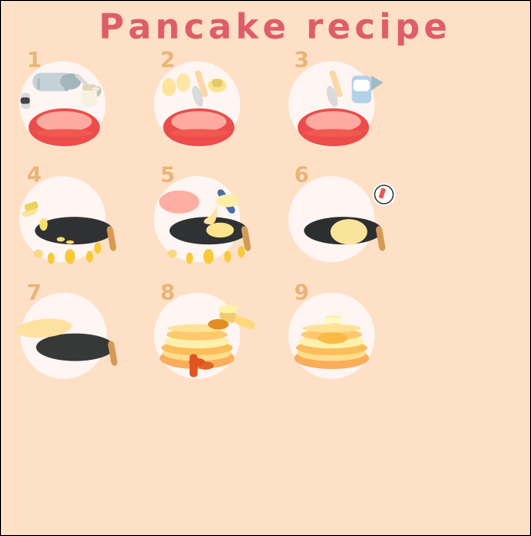}\\(d) GPT-4.1\end{minipage}
\end{center}
\begin{center}
  \textbf{Question:} What should we add in the third step? \textbf{Answer:} milk
\end{center}

\noindent
\begin{center}
  \begin{minipage}[b]{0.24\textwidth}\centering\includegraphics[width=\linewidth]{figures/mmvet_part2/25.png}\\(a) Original image\end{minipage}\hfill
  \begin{minipage}[b]{0.24\textwidth}\centering\includegraphics[width=\linewidth]{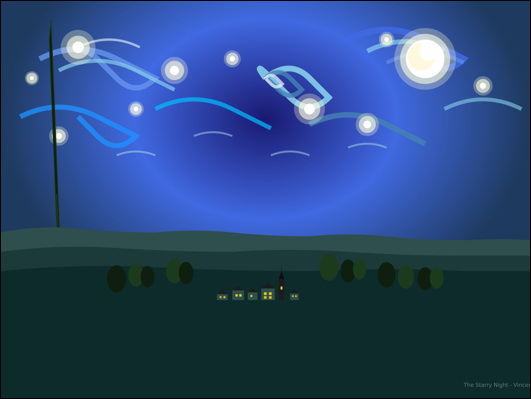}\\(b) VCoder\end{minipage}\hfill
  \begin{minipage}[b]{0.24\textwidth}\centering\includegraphics[width=\linewidth]{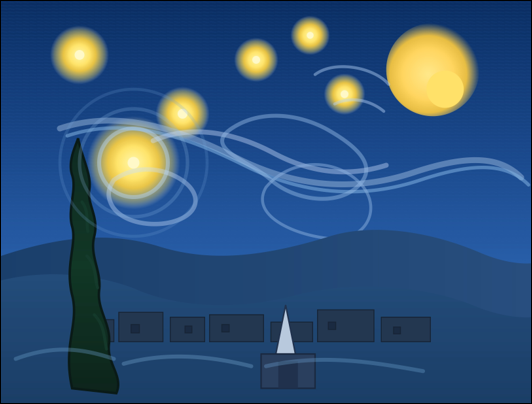}\\(c) GPT-5\end{minipage}\hfill
  \begin{minipage}[b]{0.24\textwidth}\centering\includegraphics[width=\linewidth]{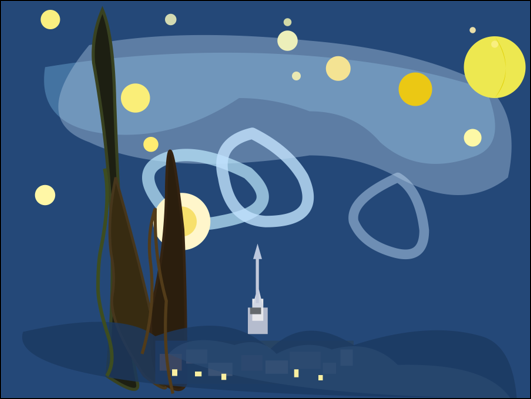}\\(d) GPT-4.1\end{minipage}
\end{center}
\begin{center}
  \textbf{Question:} Can you give a short introduction to this painting? \\
  \textbf{Answer:} The Starry Night is an oil-on-canvas painting by the Dutch Post-Impressionist painter Vincent van Gogh. Painted in June 1889, it depicts the view from the east-facing window of his asylum room at Saint-Rémy-de-Provence, just before sunrise, with the addition of an imaginary village.It has been in the permanent collection of the Museum of Modern Art in New York City since 1941, acquired through the Lillie P. Bliss Bequest. Widely regarded as Van Gogh's magnum opus, The Starry Night is one of the most recognizable paintings in Western art.
\end{center}

\noindent
\begin{center}
  \begin{minipage}[b]{0.24\textwidth}\centering\includegraphics[width=\linewidth]{figures/mmvet_part2/29.png}\\(a) Original image\end{minipage}\hfill
  \begin{minipage}[b]{0.24\textwidth}\centering\includegraphics[width=\linewidth]{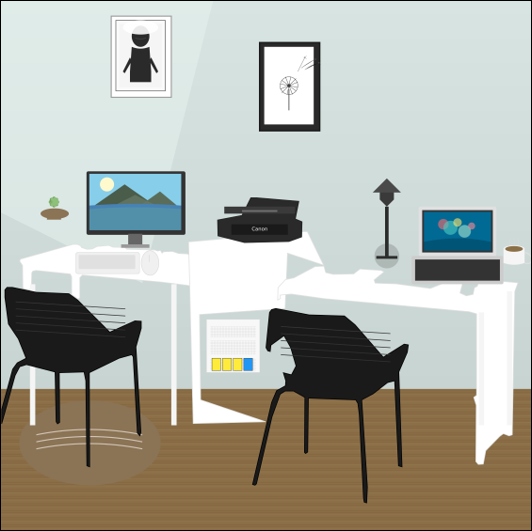}\\(b) VCoder\end{minipage}\hfill
  \begin{minipage}[b]{0.24\textwidth}\centering\includegraphics[width=\linewidth]{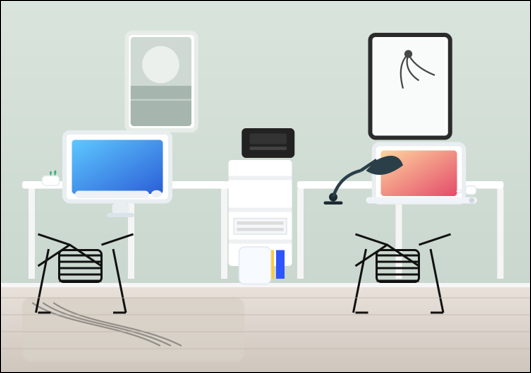}\\(c) GPT-5\end{minipage}\hfill
  \begin{minipage}[b]{0.24\textwidth}\centering\includegraphics[width=\linewidth]{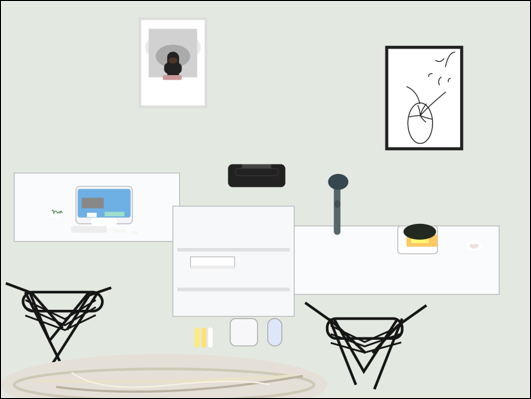}\\(d) GPT-4.1\end{minipage}
\end{center}
\begin{center}
  \textbf{Question:} On the right desk, what is to the left of the laptop?      
  \textbf{Answer:} table lamp/desk lamp
\end{center}

\noindent
\begin{center}
  \begin{minipage}[b]{0.24\textwidth}\centering\includegraphics[width=\linewidth]{figures/mmvet_part2/1.png}\\(a) Original image\end{minipage}\hfill
  \begin{minipage}[b]{0.24\textwidth}\centering\includegraphics[width=\linewidth]{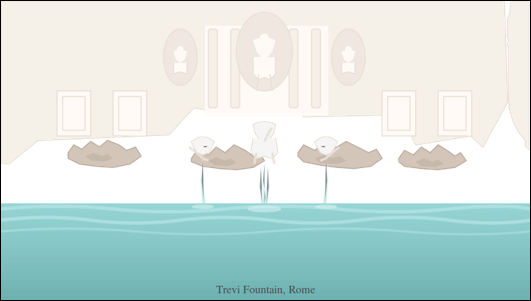}\\(b) VCoder\end{minipage}\hfill
  \begin{minipage}[b]{0.24\textwidth}\centering\includegraphics[width=\linewidth]{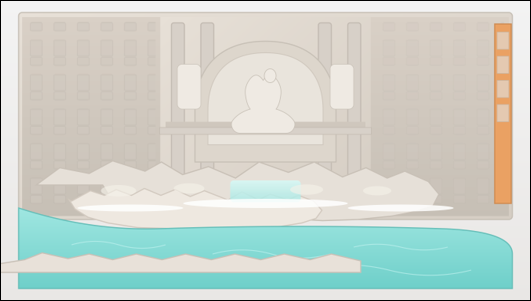}\\(c) GPT-5\end{minipage}\hfill
  \begin{minipage}[b]{0.24\textwidth}\centering\includegraphics[width=\linewidth]{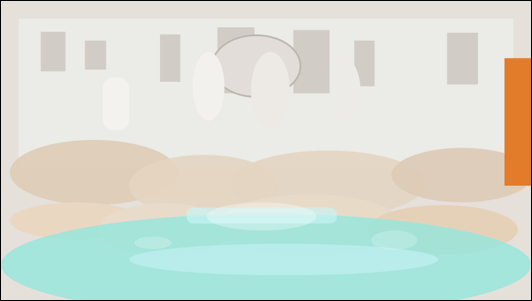}\\(d) GPT-4.1\end{minipage}
\end{center}
\begin{center}
  \textbf{Question:} What is the name of this landmark? \textbf{Answer:} Trevi Fountain
\end{center}

\noindent
\begin{center}
  \begin{minipage}[b]{0.24\textwidth}\centering\includegraphics[width=\linewidth]{figures/mmvet_part2/5.png}\\(a) Original image\end{minipage}\hfill
  \begin{minipage}[b]{0.24\textwidth}\centering\includegraphics[width=\linewidth]{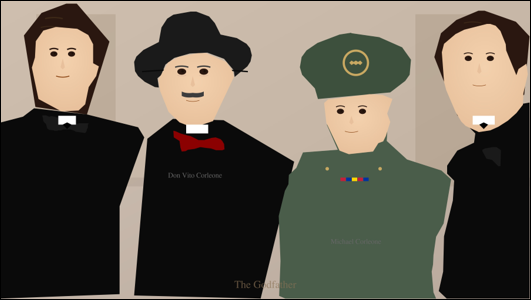}\\(b) VCoder\end{minipage}\hfill
  \begin{minipage}[b]{0.24\textwidth}\centering\includegraphics[width=\linewidth]{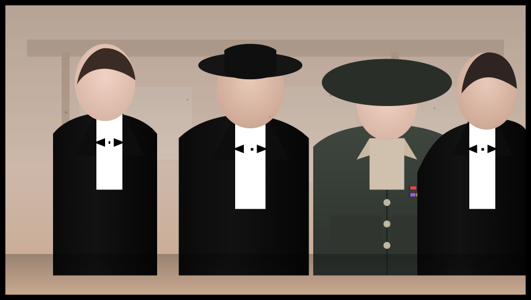}\\(c) GPT-5\end{minipage}\hfill
  \begin{minipage}[b]{0.24\textwidth}\centering\includegraphics[width=\linewidth]{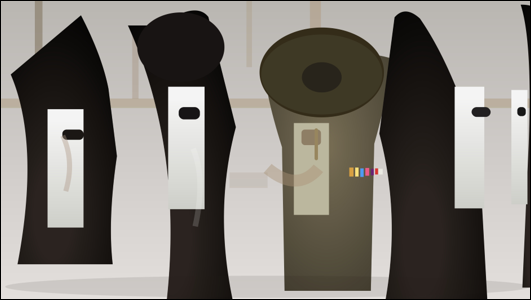}\\(d) GPT-4.1\end{minipage}
\end{center}
\begin{center}
  \textbf{Question:} Can you give a short introduction to this movie? \\
  \textbf{Answer:} The Godfather is a 1972 American crime film[2] directed by Francis Ford Coppola, who co-wrote the screenplay with Mario Puzo, based on Puzo's best-selling 1969 novel of the same title. The film stars Marlon Brando, Al Pacino, James Caan, Richard Castellano, Robert Duvall, Sterling Hayden, John Marley, Richard Conte, and Diane Keaton. It is the first installment in The Godfather trilogy, chronicling the Corleone family under patriarch Vito Corleone (Brando) from 1945 to 1955. It focuses on the transformation of his youngest son, Michael Corleone (Pacino), from reluctant family outsider to ruthless mafia boss.
\end{center}

\noindent
\begin{center}
  \begin{minipage}[b]{0.24\textwidth}\centering\includegraphics[width=\linewidth]{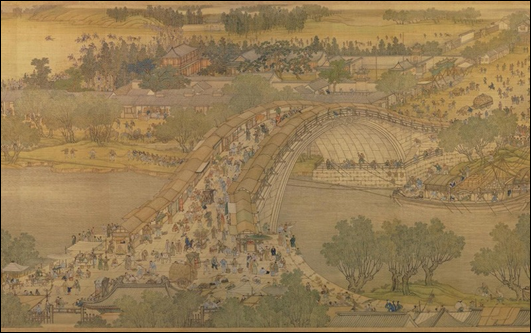}\\(a) Original image\end{minipage}\hfill
  \begin{minipage}[b]{0.24\textwidth}\centering\includegraphics[width=\linewidth]{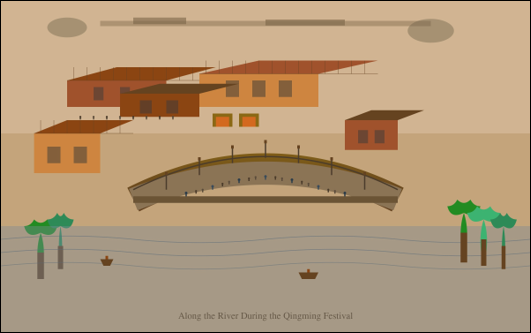}\\(b) VCoder\end{minipage}\hfill
  \begin{minipage}[b]{0.24\textwidth}\centering\includegraphics[width=\linewidth]{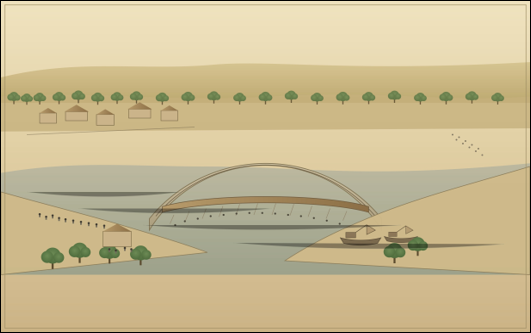}\\(c) GPT-5\end{minipage}\hfill
  \begin{minipage}[b]{0.24\textwidth}\centering\includegraphics[width=\linewidth]{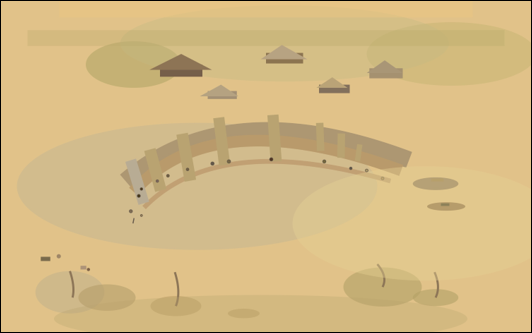}\\(d) GPT-4.1\end{minipage}
\end{center}
\begin{center}
  \textbf{Question:} Can you give a short introduction to this painting? \\
  \textbf{Answer:} Along the River During the Qingming Festival (Qingming Shanghe Tu) is a handscroll painting by the Song dynasty painter Zhang Zeduan (1085–1145) and copied many times in the following centuries. It captures the daily life of people and the landscape of the capital, Bianjing (present-day Kaifeng) during the Northern Song. The theme is often said to celebrate the festive spirit and worldly commotion at the Qingming Festival, rather than the holiday's ceremonial aspects, such as tomb sweeping and prayers. Read right to left, as a viewer unrolled it, successive scenes reveal the lifestyle of all levels of the society from rich to poor as well as economic activities in rural areas and the city, and offer glimpses of period clothing and architecture. The painting is considered to be the most renowned work among all Chinese paintings, and it has been called "China's Mona Lisa."
\end{center}

\noindent
\begin{center}
  \begin{minipage}[b]{0.24\textwidth}\centering\includegraphics[width=\linewidth]{figures/mmvet_part2/13.png}\\(a) Original image\end{minipage}\hfill
  \begin{minipage}[b]{0.24\textwidth}\centering\includegraphics[width=\linewidth]{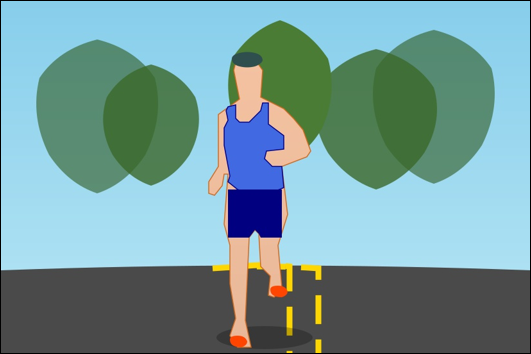}\\(b) VCoder\end{minipage}\hfill
  \begin{minipage}[b]{0.24\textwidth}\centering\includegraphics[width=\linewidth]{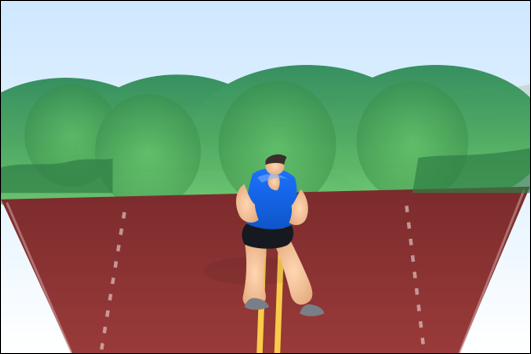}\\(c) GPT-5\end{minipage}\hfill
  \begin{minipage}[b]{0.24\textwidth}\centering\includegraphics[width=\linewidth]{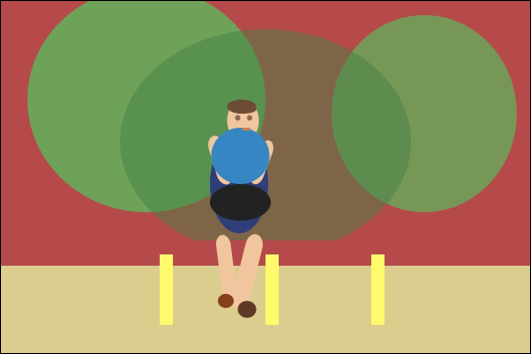}\\(d) GPT-4.1\end{minipage}
\end{center}
\begin{center}
  \textbf{Question:} Is the man going to fall down? \textbf{Answer:} no
\end{center}
\subsubsection{MMMU}
\noindent
\begin{center}
  \begin{minipage}[b]{0.48\textwidth}\centering\includegraphics[width=\linewidth]{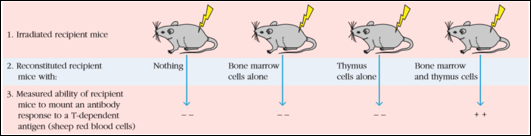}\\(a) Original image\end{minipage}\hfill
  \begin{minipage}[b]{0.48\textwidth}\centering\includegraphics[width=\linewidth]{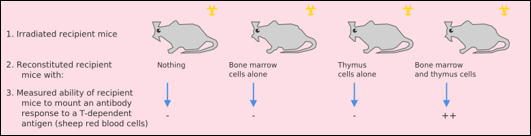}\\(b) VCoder\end{minipage}\hfill
  
  \begin{minipage}[b]{0.48\textwidth}\centering\includegraphics[width=\linewidth]{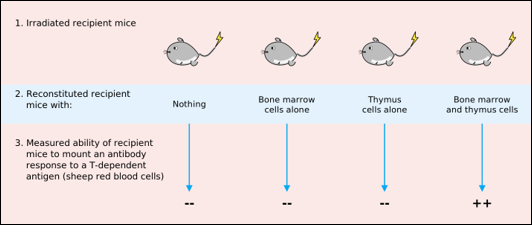}\\(c) Gemini-2.5-Pro\end{minipage}\hfill
  \begin{minipage}[b]{0.48\textwidth}\centering\includegraphics[width=\linewidth]{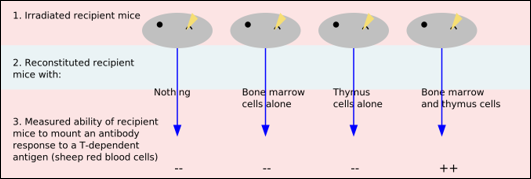}\\(d) GPT-4o\end{minipage}
\end{center}
\begin{center}
  \textbf{Question:} For your independent research, you transferred lymphocyte populations between syngeneic mice. You irradiated recipients first to ablate (get rid of) existing lymphocytes, then transferred defined cell populations from donors of same genetic background. The result is shown in . What does this experiment tell us?
(A) Both B cells and T cells can produce antibodies.
(B) Both B cells and T cells have memory functions.
(C) Both B cells and T cells are required for an antibody response.
(D) B cells are required for an antibody response in the absence of T cells.
(E) B cells and T cells are co-localized and produce synergetic effects in bone marrow and thymus.\\
Answer with the option's letter from the given choices directly.\\
\textbf{Answer:} C
\end{center}

\noindent
\begin{center}
  \begin{minipage}[b]{0.48\textwidth}\centering\includegraphics[width=\linewidth]{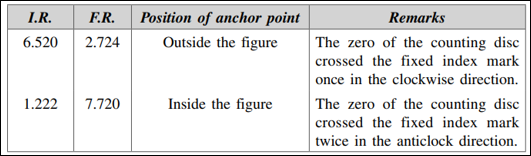}\\(a) Original image\end{minipage}\hfill
  \begin{minipage}[b]{0.48\textwidth}\centering\includegraphics[width=\linewidth]{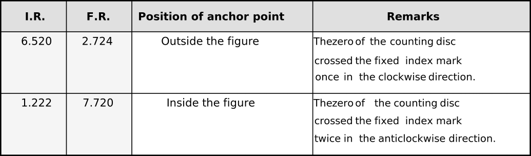}\\(b) VCoder\end{minipage}\hfill
  
  \begin{minipage}[b]{0.48\textwidth}\centering\includegraphics[width=\linewidth]{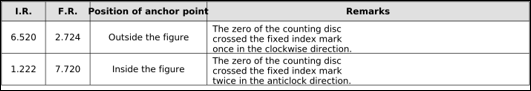}\\(c) Gemini-2.5-Pro\end{minipage}\hfill
  \begin{minipage}[b]{0.48\textwidth}\centering\includegraphics[width=\linewidth]{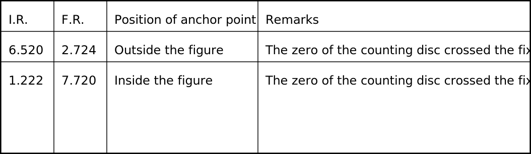}\\(d) GPT-4o\end{minipage}
\end{center}
\begin{center}
  \textbf{Question:} Calculate the area of the zero circle with the following data:Assume that the tracing arm of the planimeter was so set that one revolution of the measuring wheel measures 100 $cm^{2}$ on the paper.
Answer the question using a single word or phrase. \\
\textbf{Answer:} 1970.6
\end{center}

\noindent
\begin{center}
  \begin{minipage}[b]{0.24\textwidth}\centering\includegraphics[width=\linewidth]{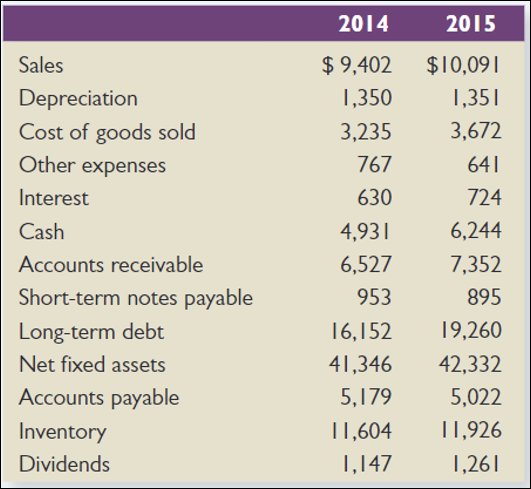}\\(a) Original image\end{minipage}\hfill
  \begin{minipage}[b]{0.24\textwidth}\centering\includegraphics[width=\linewidth]{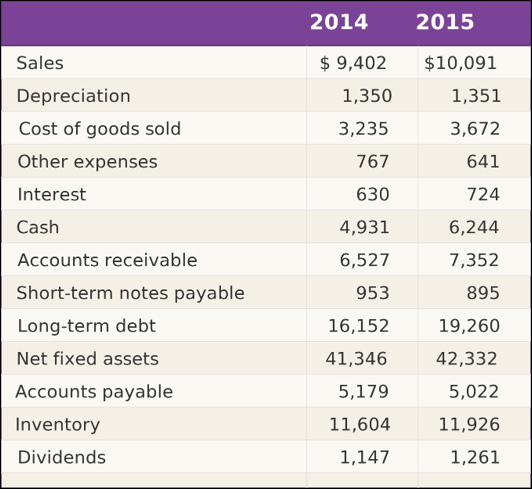}\\(b) VCoder\end{minipage}\hfill
  \begin{minipage}[b]{0.24\textwidth}\centering\includegraphics[width=\linewidth]{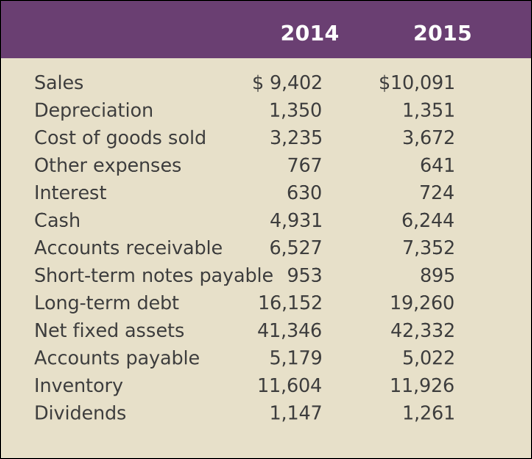}\\(c) Gemini-2.5-Pro\end{minipage}\hfill
  \begin{minipage}[b]{0.24\textwidth}\centering\includegraphics[width=\linewidth]{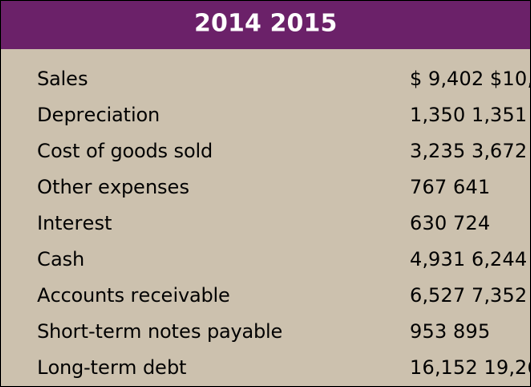}\\(d) GPT-4o\end{minipage}
\end{center}
\begin{center}
  \textbf{Question:} For 2015, calculate the cash flow from assets(1) \_\_\_\_\_, 
  cash flow to creditors(2) \_\_\_\_\_, and cash flow to stockholders(3) \_\_\_\_\_. (A) 1): -\textdollar493.02 (2):-\textdollar2,384 (3):\textdollar1,890.98 (B) 1): \textdollar1843.98 (2):-\textdollar2,384 (3):\textdollar493.02 (C) 1): -\textdollar493.02 (2):-\textdollar2,384 (3):-\textdollar1,890.98\\
  Answer with the option’s letter from the given choices directly.\\
  \textbf{Answer:} C
\end{center}

\noindent
\begin{center}
  \begin{minipage}[b]{0.24\textwidth}\centering\includegraphics[width=\linewidth]{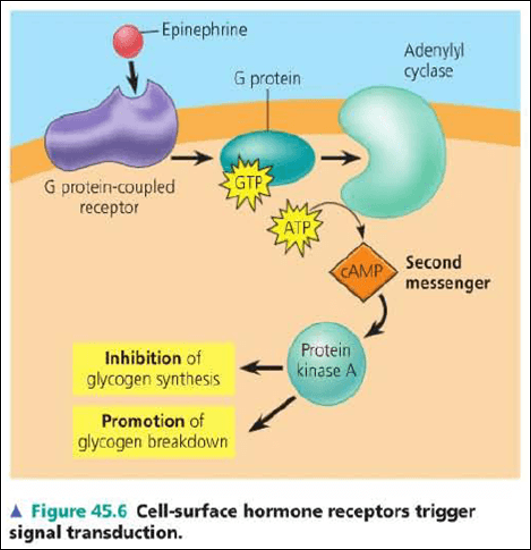}\\(a) Original image\end{minipage}\hfill
  \begin{minipage}[b]{0.24\textwidth}\centering\includegraphics[width=\linewidth]{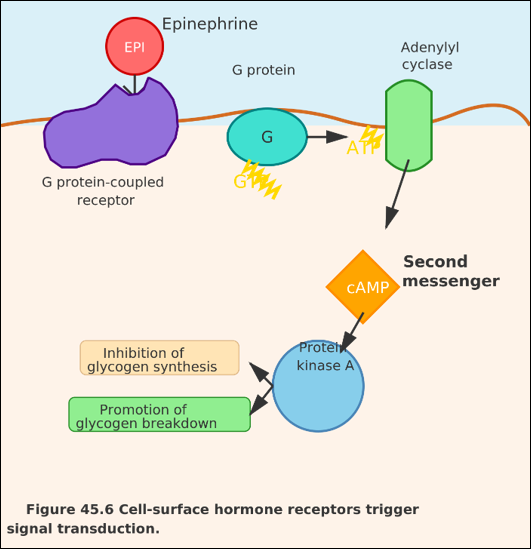}\\(b) VCoder\end{minipage}\hfill
  \begin{minipage}[b]{0.24\textwidth}\centering\includegraphics[width=\linewidth]{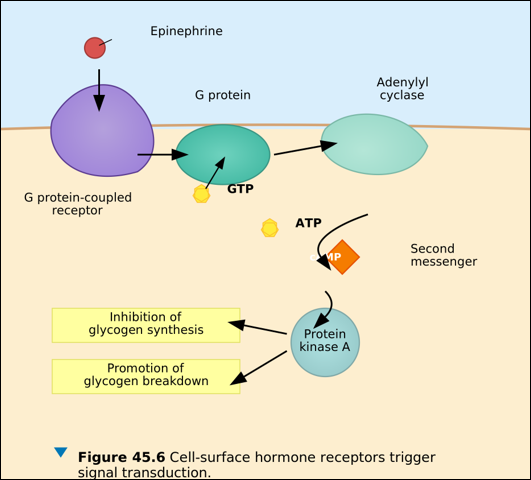}\\(c) Gemini-2.5-Pro\end{minipage}\hfill
  \begin{minipage}[b]{0.24\textwidth}\centering\includegraphics[width=\linewidth]{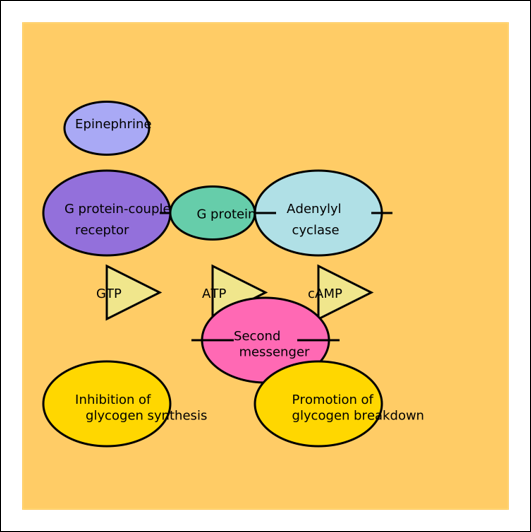}\\(d) GPT-4o\end{minipage}
\end{center}
\begin{center}
\textbf{Question:} Which of the following correctly describes the reception stage of this signal transduction pathway?
(A) epinephrine binds to a g-protein coupled receptor protein present in the cell membrane
(B) the g protein changes shape, is activated, activates adenyl cyclase, which activates cAMP, which activates protein kinases
(C) protein kinases phosphoylate molecules
(D) glycogen synthesis is inhibited and glycogen breakdown is promoted\\
Answer with the option's letter from the given choices directly. \\
\textbf{Answer:} A
\end{center}

\noindent
\begin{center}
  \begin{minipage}[b]{0.24\textwidth}\centering\includegraphics[width=\linewidth]{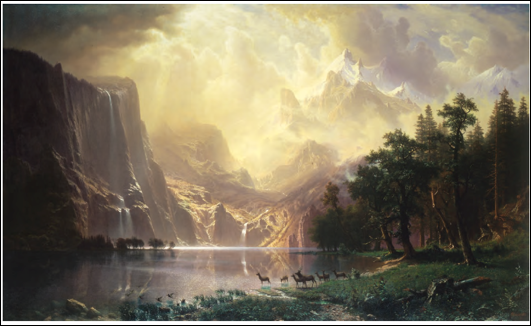}\\(a) Original image\end{minipage}\hfill
  \begin{minipage}[b]{0.24\textwidth}\centering\includegraphics[width=\linewidth]{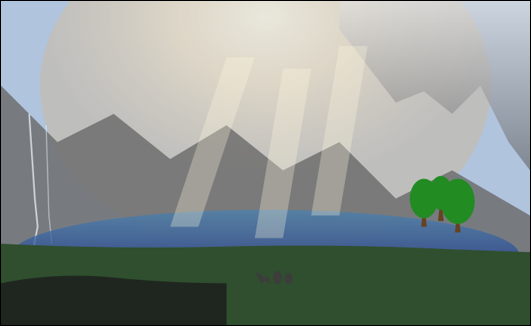}\\(b) VCoder\end{minipage}\hfill
  \begin{minipage}[b]{0.24\textwidth}\centering\includegraphics[width=\linewidth]{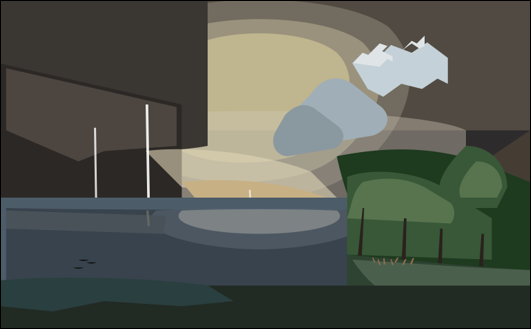}\\(c) Gemini-2.5-Pro\end{minipage}\hfill
  \begin{minipage}[b]{0.24\textwidth}\centering\includegraphics[width=\linewidth]{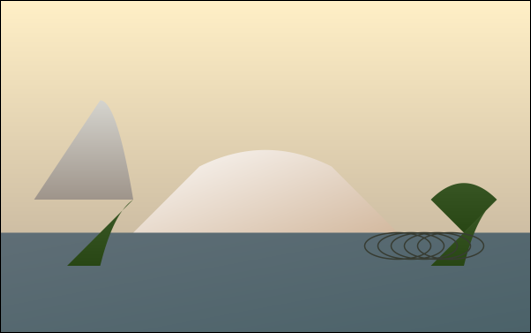}\\(d) GPT-4o\end{minipage}
\end{center}
\begin{center}
  \textbf{Question:} The painting on the right focuses on the 
(A) contribution of Native Americans to landscape preservation
(B) implementation of the Homestead Act
(C) impact of the gold rush on landscape development
(D) idea of Manifest Destiny\\
Answer with the option's letter from the given choices directly. \\ 
\textbf{Answer:} D
\end{center}

\noindent
\begin{center}
  \begin{minipage}[b]{0.24\textwidth}\centering\includegraphics[width=\linewidth]{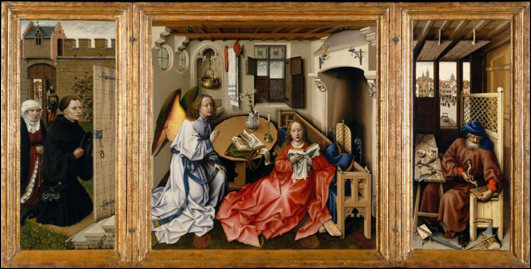}\\(a) Original image\end{minipage}\hfill
  \begin{minipage}[b]{0.24\textwidth}\centering\includegraphics[width=\linewidth]{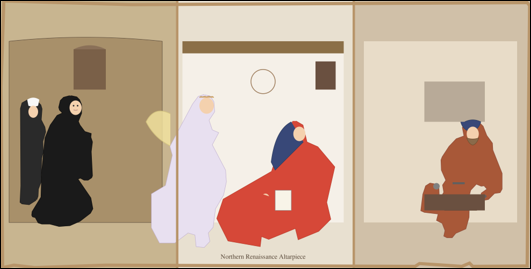}\\(b) VCoder\end{minipage}\hfill
  \begin{minipage}[b]{0.24\textwidth}\centering\includegraphics[width=\linewidth]{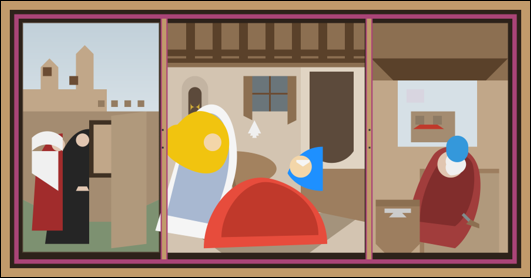}\\(c) Gemini-2.5-Pro\end{minipage}\hfill
  \begin{minipage}[b]{0.24\textwidth}\centering\includegraphics[width=\linewidth]{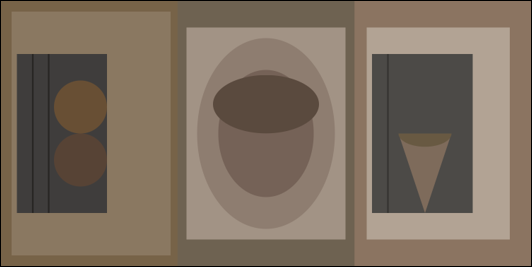}\\(d) GPT-4o\end{minipage}
\end{center}
\begin{center}
  \textbf{Question:} Both works come from which art-historical period?
(A) Baroque
(B) Renaissance
(C) Rococo
(D) Classical 
Answer with the option's letter from the given choices directly.\\
\textbf{Answer:} B
\end{center}

\noindent
\begin{center}
  \begin{minipage}[b]{0.24\textwidth}\centering\includegraphics[width=\linewidth]{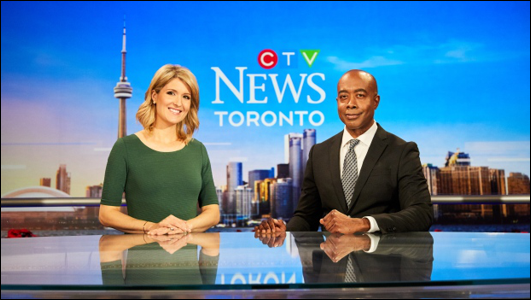}\\(a) Original image\end{minipage}\hfill
  \begin{minipage}[b]{0.24\textwidth}\centering\includegraphics[width=\linewidth]{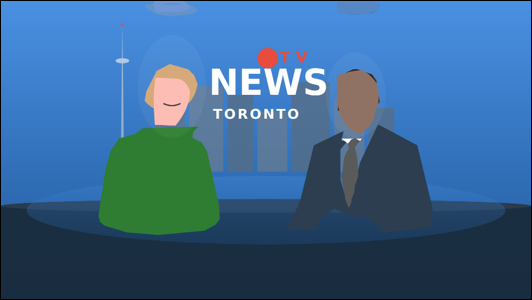}\\(b) VCoder\end{minipage}\hfill
  \begin{minipage}[b]{0.24\textwidth}\centering\includegraphics[width=\linewidth]{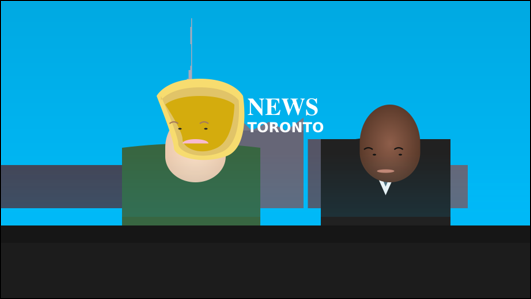}\\(c) Gemini-2.5-Pro\end{minipage}\hfill
  \begin{minipage}[b]{0.24\textwidth}\centering\includegraphics[width=\linewidth]{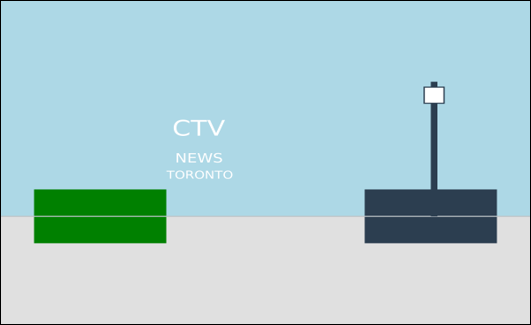}\\(d) GPT-4o\end{minipage}
\end{center}
\begin{center}
  \textbf{Question:} Refer to the figure, which term best describes the practice where students take on the role of television or newspaper reporters and interview characters from the book to retell an event from a range of perspectives?
(A) News Program
(B) Readers Theatre
(C) Hot Seat
(D) News\\
Answer with the option's letter from the given choices directly. \\
\textbf{Answer:} A
\end{center}

\noindent
\begin{center}
  \begin{minipage}[b]{0.24\textwidth}\centering\includegraphics[width=\linewidth]{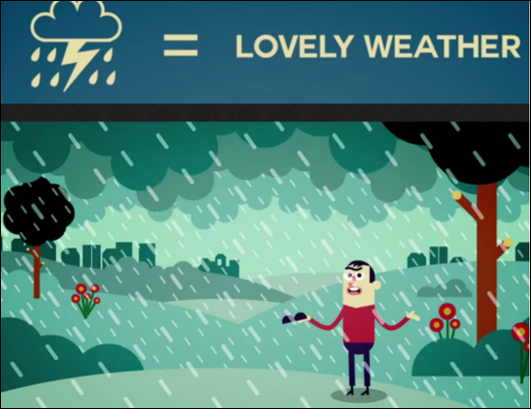}\\(a) Original image\end{minipage}\hfill
  \begin{minipage}[b]{0.24\textwidth}\centering\includegraphics[width=\linewidth]{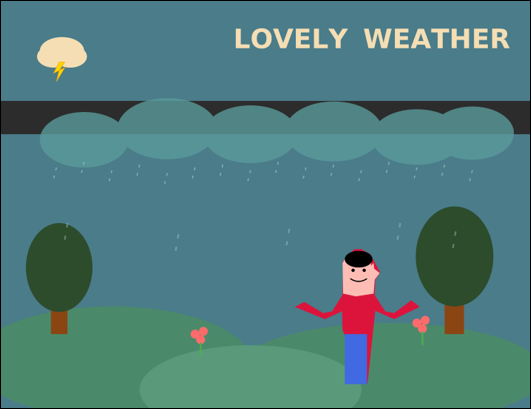}\\(b) VCoder\end{minipage}\hfill
  \begin{minipage}[b]{0.24\textwidth}\centering\includegraphics[width=\linewidth]{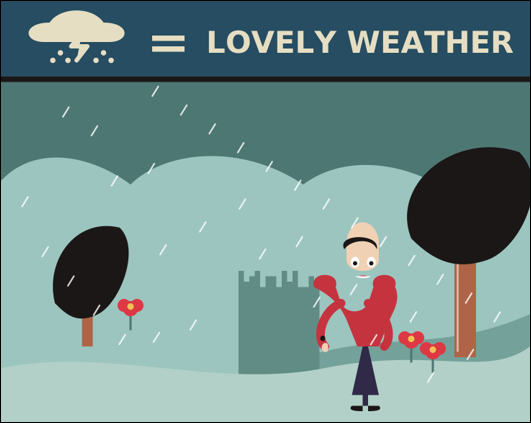}\\(c) Gemini-2.5-Pro\end{minipage}\hfill
  \begin{minipage}[b]{0.24\textwidth}\centering\includegraphics[width=\linewidth]{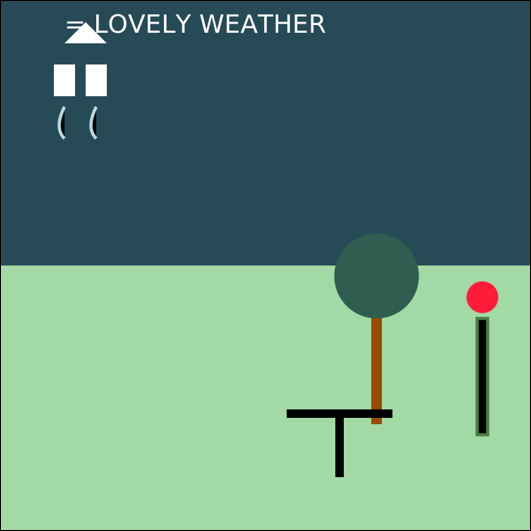}\\(d) GPT-4o\end{minipage}
\end{center}
\begin{center}
  \textbf{Question:} Refer to the description , which type of irony is depicted when a person says or writes one thing and means another, or uses words to convey a meaning opposite to the literal meaning?
(A) verbal irony
(B) situational irony
(C) foreshadowing
(D) dramatic irony\\
Answer with the option's letter from the given choices directly.\\
\textbf{Answer:} A
\end{center}
\subsubsection{CV-Bench}
\noindent
\begin{center}
  \begin{minipage}[b]{0.24\textwidth}\centering\includegraphics[width=\linewidth]{figures/cv-bench/1.png}\\(a) Original image\end{minipage}\hfill
  \begin{minipage}[b]{0.24\textwidth}\centering\includegraphics[width=\linewidth]{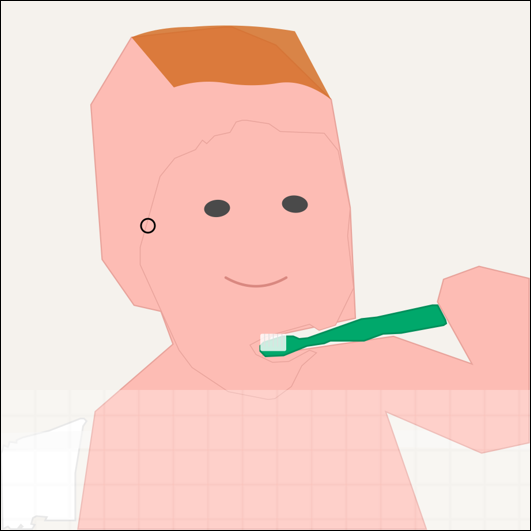}\\(b) VCoder\end{minipage}\hfill
  \begin{minipage}[b]{0.24\textwidth}\centering\includegraphics[width=\linewidth]{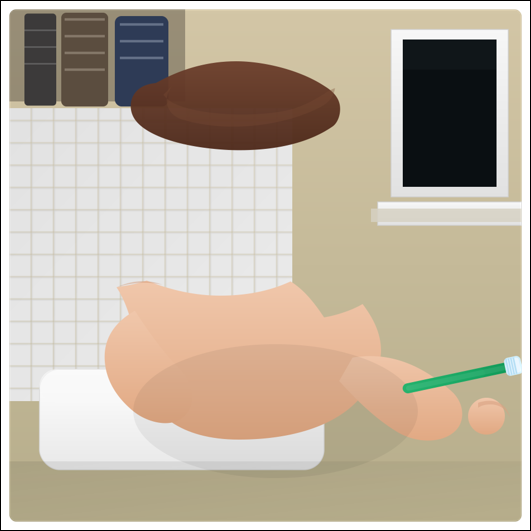}\\(c) GPT-5\end{minipage}\hfill
  \begin{minipage}[b]{0.24\textwidth}\centering\includegraphics[width=\linewidth]{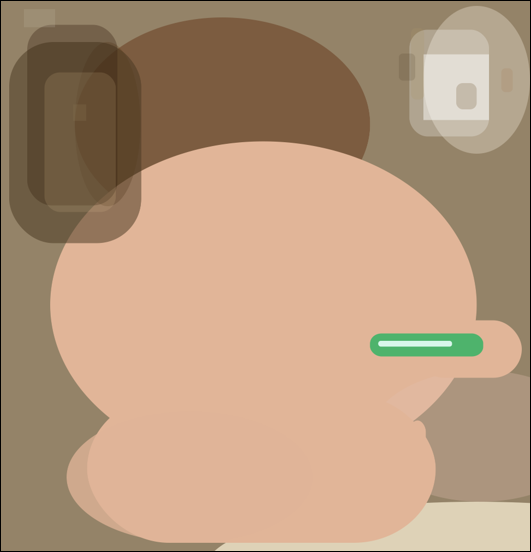}\\(d) GPT-4.1\end{minipage}
\end{center}
\begin{center}
\textbf{Question:} How many persons are in the image? Select from the following choices.
(A) 2
(B) 3
(C) 0
(D) 1
Answer with the option's letter from the given choices directly.\\
 \textbf{Answer:} D
\end{center}

\noindent
\begin{center}
  \begin{minipage}[b]{0.24\textwidth}\centering\includegraphics[width=\linewidth]{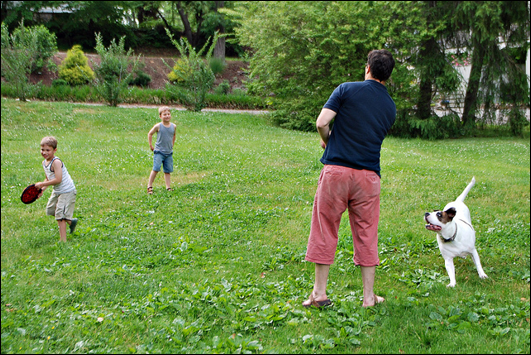}\\(a) Original image\end{minipage}\hfill
  \begin{minipage}[b]{0.24\textwidth}\centering\includegraphics[width=\linewidth]{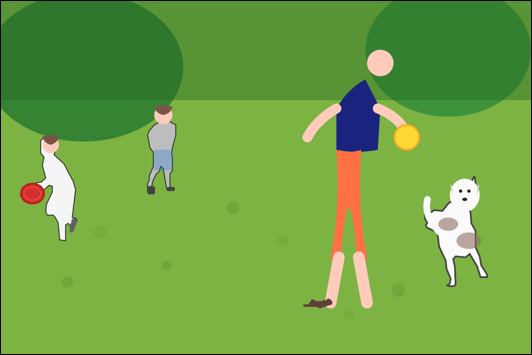}\\(b) VCoder\end{minipage}\hfill
  \begin{minipage}[b]{0.24\textwidth}\centering\includegraphics[width=\linewidth]{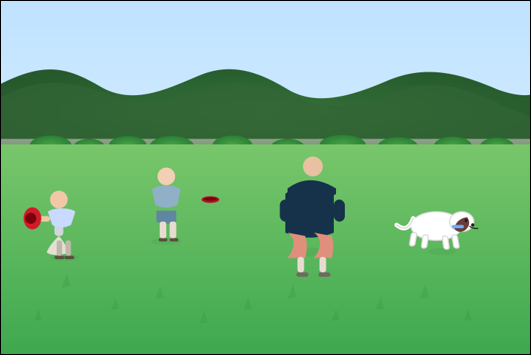}\\(c) GPT-5\end{minipage}\hfill
  \begin{minipage}[b]{0.24\textwidth}\centering\includegraphics[width=\linewidth]{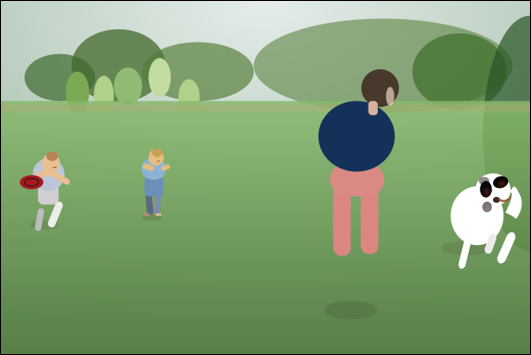}\\(d) GPT-4.1\end{minipage}
\end{center}
\begin{center}
  \textbf{Question:} How many dogs are in the image? Select from the following choices.
(A) 1
(B) 3
(C) 2
(D) 0
Answer with the option's letter from the given choices directly. \\ 
\textbf{Answer:} A
\end{center}

\noindent
\begin{center}
  \begin{minipage}[b]{0.24\textwidth}\centering\includegraphics[width=\linewidth]{figures/cv-bench/9.png}\\(a) Original image\end{minipage}\hfill
  \begin{minipage}[b]{0.24\textwidth}\centering\includegraphics[width=\linewidth]{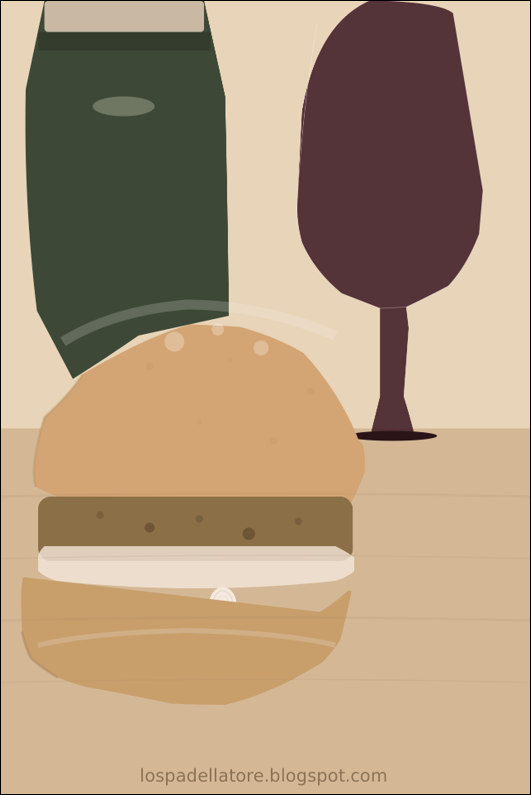}\\(b) VCoder\end{minipage}\hfill
  \begin{minipage}[b]{0.24\textwidth}\centering\includegraphics[width=\linewidth]{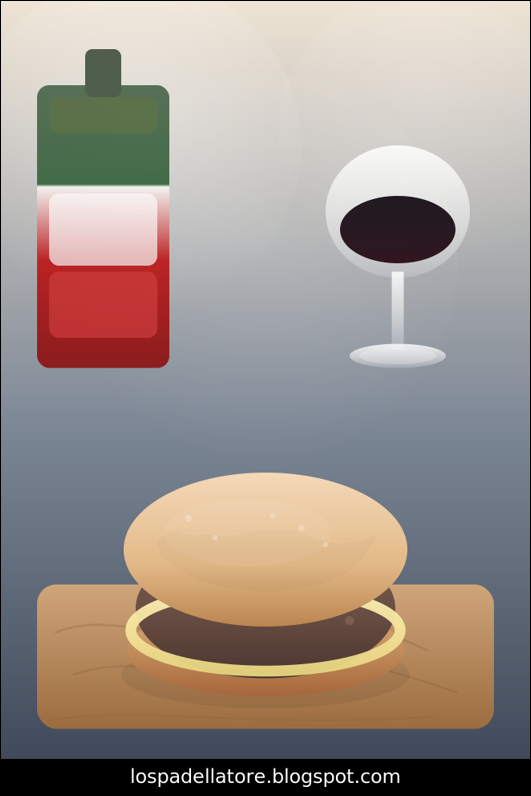}\\(c) GPT-5\end{minipage}\hfill
  \begin{minipage}[b]{0.24\textwidth}\centering\includegraphics[width=\linewidth]{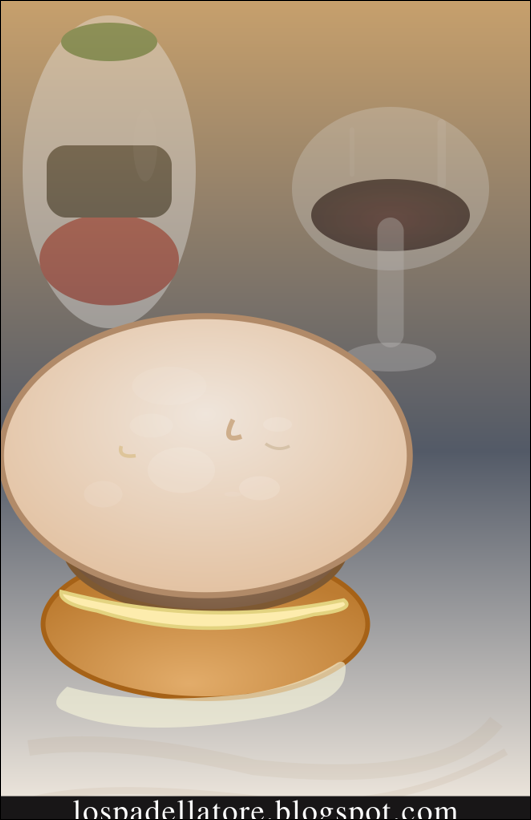}\\(d) GPT-4.1\end{minipage}
\end{center}
\begin{center}
  \textbf{Question:} Considering the relative positions of the bottle and the wine glass in the image provided, where is the bottle located with respect to the wine glass? Select from the following choices.
(A) left
(B) right  Answer with the option's letter from the given choices directly. \\
\textbf{Answer:} A
\end{center}

\noindent
\begin{center}
  \begin{minipage}[b]{0.24\textwidth}\centering\includegraphics[width=\linewidth]{figures/cv-bench/13.png}\\(a) Original image\end{minipage}\hfill
  \begin{minipage}[b]{0.24\textwidth}\centering\includegraphics[width=\linewidth]{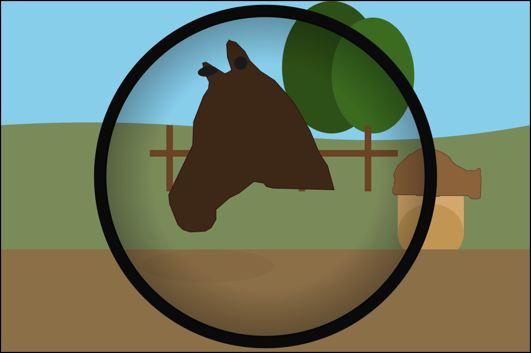}\\(b) VCoder\end{minipage}\hfill
  \begin{minipage}[b]{0.24\textwidth}\centering\includegraphics[width=\linewidth]{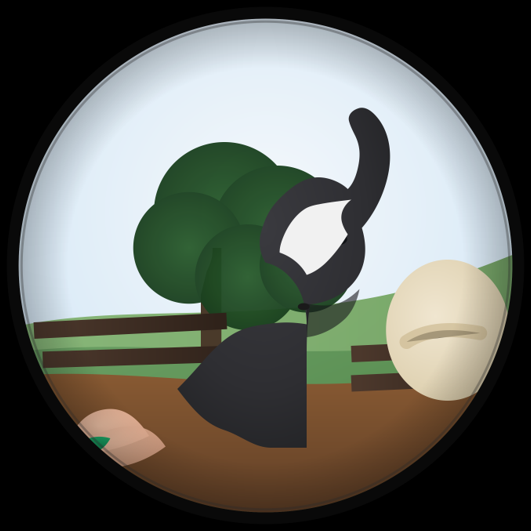}\\(c) GPT-5\end{minipage}\hfill
  \begin{minipage}[b]{0.24\textwidth}\centering\includegraphics[width=\linewidth]{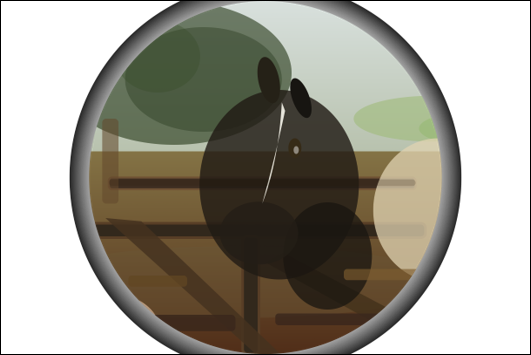}\\(d) GPT-4.1\end{minipage}
\end{center}
\begin{center}
  \textbf{Question:} Considering the relative positions of the sheep and the horse in the image provided, where is the sheep located with respect to the horse? Select from 
(A) left
(B) right \\
Answer with the option's letter from the given choices directly.\\
\textbf{Answer:} B
\end{center}

\noindent
\begin{center}
  \begin{minipage}[b]{0.24\textwidth}\centering\includegraphics[width=\linewidth]{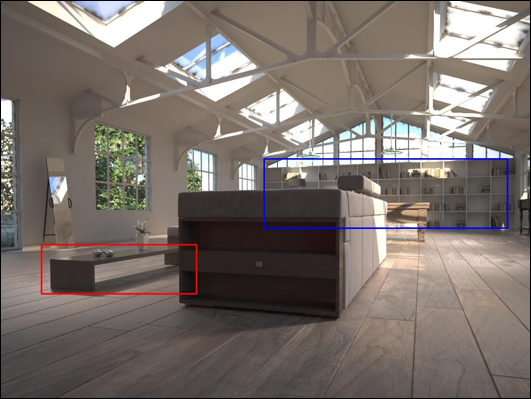}\\(a) Original image\end{minipage}\hfill
  \begin{minipage}[b]{0.24\textwidth}\centering\includegraphics[width=\linewidth]{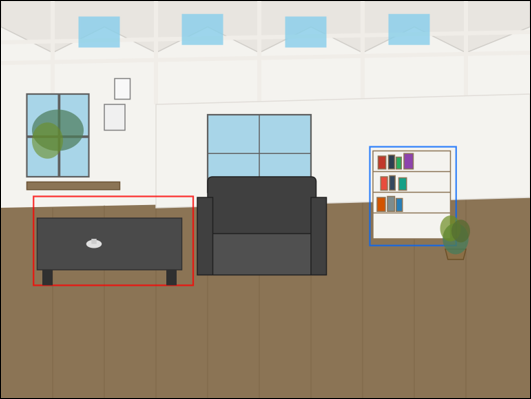}\\(b) VCoder\end{minipage}\hfill
  \begin{minipage}[b]{0.24\textwidth}\centering\includegraphics[width=\linewidth]{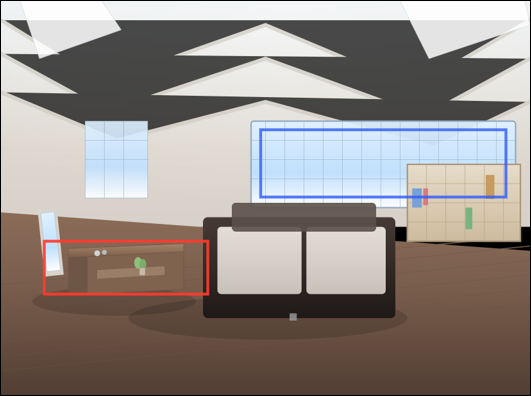}\\(c) GPT-5\end{minipage}\hfill
  \begin{minipage}[b]{0.24\textwidth}\centering\includegraphics[width=\linewidth]{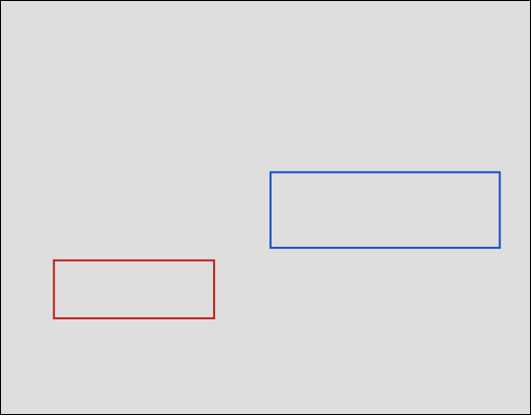}\\(d) GPT-4.1\end{minipage}
\end{center}
\begin{center}
  \textbf{Question:} Which object is closer to the camera taking this photo, the table (highlighted by a red box) or the bookcase (highlighted by a blue box)?
(A) table
(B) bookcase \\
Answer with the option's letter from the given choices directly. \\
\textbf{Answer:} A
\end{center}

\noindent
\begin{center}
  \begin{minipage}[b]{0.24\textwidth}\centering\includegraphics[width=\linewidth]{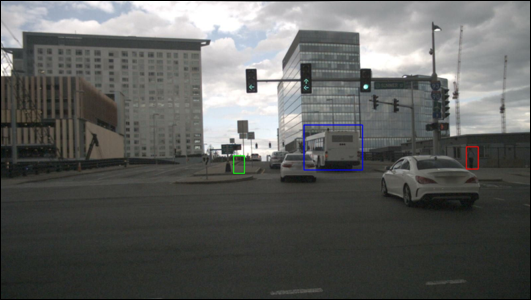}\\(a) Original image\end{minipage}\hfill
  \begin{minipage}[b]{0.24\textwidth}\centering\includegraphics[width=\linewidth]{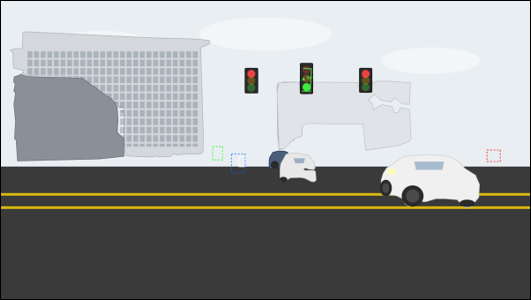}\\(b) VCoder\end{minipage}\hfill
  \begin{minipage}[b]{0.24\textwidth}\centering\includegraphics[width=\linewidth]{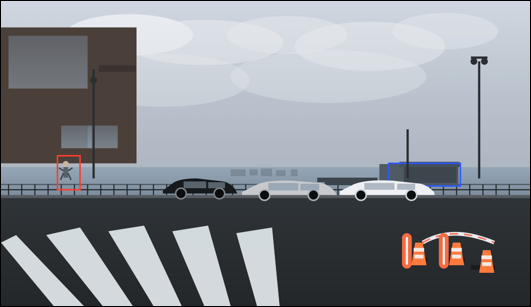}\\(c) GPT-5\end{minipage}\hfill
  \begin{minipage}[b]{0.24\textwidth}\centering\includegraphics[width=\linewidth]{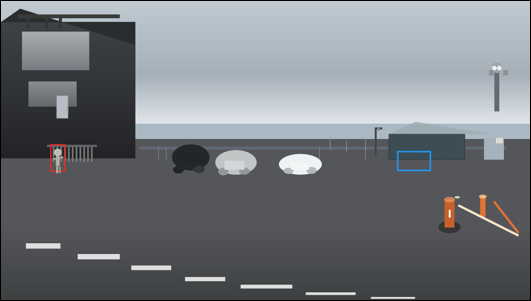}\\(d) GPT-4.1\end{minipage}
\end{center}
\begin{center}
  \textbf{Question:} Estimate the real-world distances between objects in this image. Which object is closer to the traffic cone (highlighted by a red box), the trailer (highlighted by a blue box) or the bus (highlighted by a green box)?
(A) trailer
(B) bus \\
Answer with the option's letter from the given choices directly.\\
\textbf{Answer:} A
\end{center}

\noindent
\begin{center}
  \begin{minipage}[b]{0.24\textwidth}\centering\includegraphics[width=\linewidth]{figures/cv-bench/25.png}\\(a) Original image\end{minipage}\hfill
  \begin{minipage}[b]{0.24\textwidth}\centering\includegraphics[width=\linewidth]{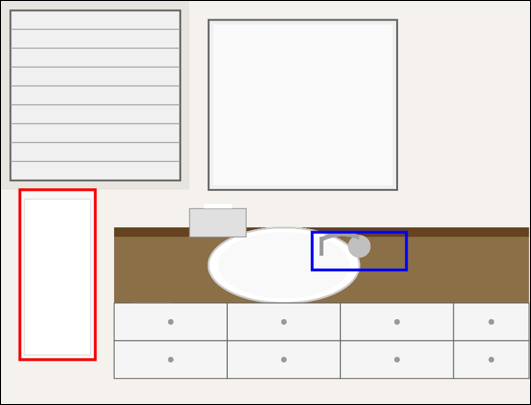}\\(b) VCoder\end{minipage}\hfill
  \begin{minipage}[b]{0.24\textwidth}\centering\includegraphics[width=\linewidth]{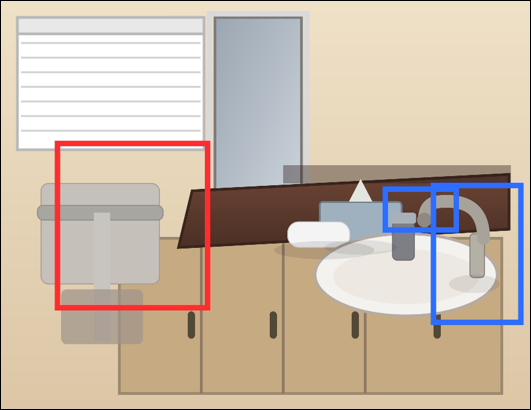}\\(c) GPT-5\end{minipage}\hfill
  \begin{minipage}[b]{0.24\textwidth}\centering\includegraphics[width=\linewidth]{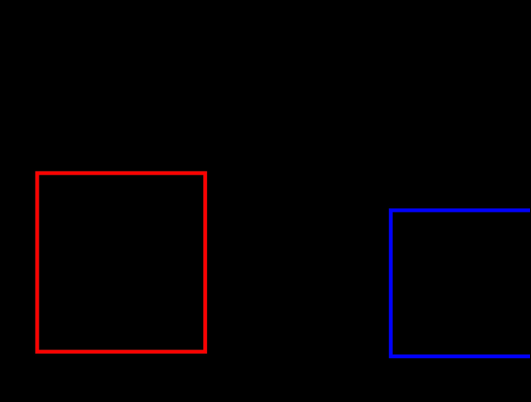}\\(d) GPT-4.1\end{minipage}
\end{center}
\begin{center}
  \textbf{Question:} Which object is closer to the camera taking this photo, the towel (highlighted by a red box) or the faucet (highlighted by a blue box)?
(A) towel
(B) faucet \\
Answer with the option's letter from the given choices directly.\\
\textbf{Answer:} B
\end{center}

\noindent
\begin{center}
  \begin{minipage}[b]{0.24\textwidth}\centering\includegraphics[width=\linewidth]{figures/cv-bench/29.png}\\(a) Original image\end{minipage}\hfill
  \begin{minipage}[b]{0.24\textwidth}\centering\includegraphics[width=\linewidth]{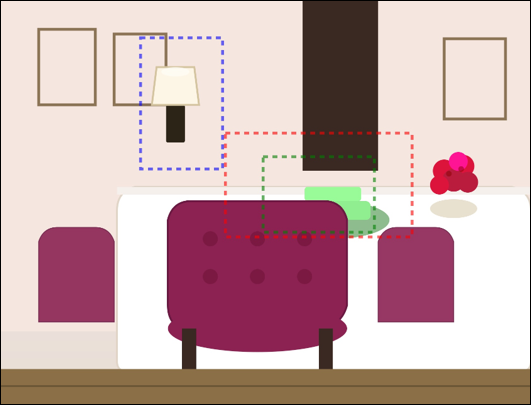}\\(b) VCoder\end{minipage}\hfill
  \begin{minipage}[b]{0.24\textwidth}\centering\includegraphics[width=\linewidth]{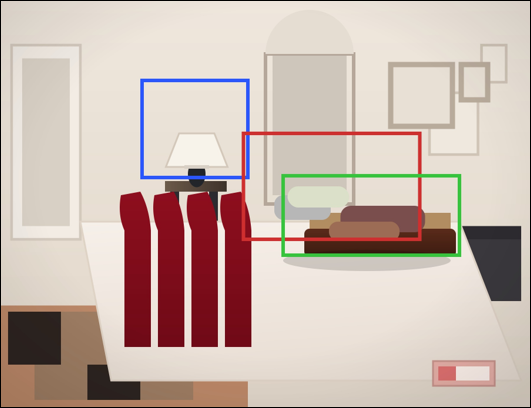}\\(c) GPT-5\end{minipage}\hfill
  \begin{minipage}[b]{0.24\textwidth}\centering\includegraphics[width=\linewidth]{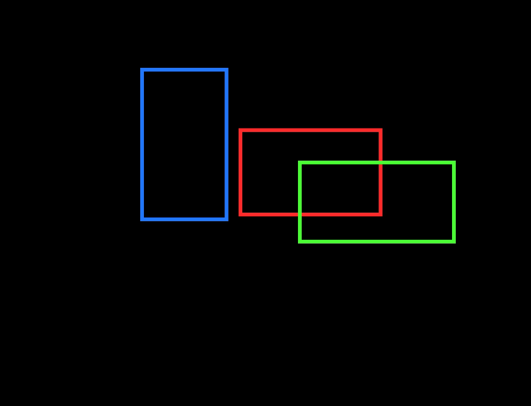}\\(d) GPT-4.1\end{minipage}
\end{center}
\begin{center}
  \textbf{Question:} Estimate the real-world distances between objects in this image. Which object is closer to the clothes (highlighted by a red box), the lamp (highlighted by a blue box) or the towel (highlighted by a green box)?
(A) lamp
(B) towel\\
Answer with the option's letter from the given choices directly. \\
\textbf{Answer:} B
\end{center}

\subsection{VCoder Individual Components}
In this section, we present ablation studies visualizing the contribution of individual components in VCoder. For each example, we show: (a) the original reference image, (b) the initial rendered output without any refinement, (c) the output after applying visual tools, and (d) the final output after the revision module. These progressive visualizations demonstrate how each component incrementally improves the quality and accuracy of the generated images.
\noindent
\begin{center}
  \begin{minipage}[b]{0.24\textwidth}\centering\includegraphics[width=\linewidth]{figures/mm-vet/1.png}\\(a) Original image\end{minipage}\hfill
  \begin{minipage}[b]{0.24\textwidth}\centering\includegraphics[width=\linewidth]{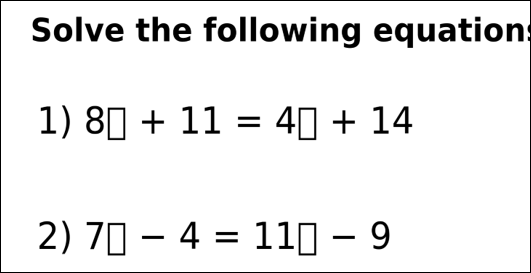}\\(b) Initial rendered\end{minipage}\hfill
  \begin{minipage}[b]{0.24\textwidth}\centering\includegraphics[width=\linewidth]{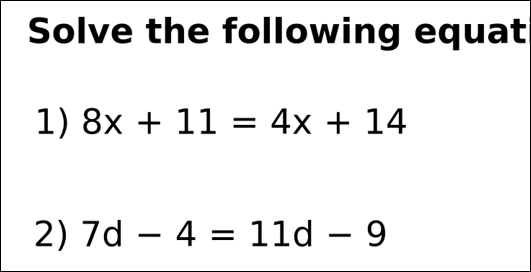}\\(c) w. visual tools\end{minipage}\hfill
  \begin{minipage}[b]{0.24\textwidth}\centering\includegraphics[width=\linewidth]{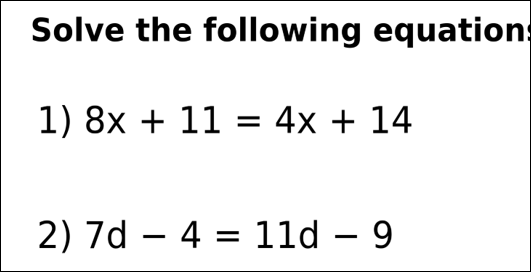}\\(d) w. revision\end{minipage}
\end{center}
\begin{center}
  \textbf{Question:} What is $d$ in the last equation? \textbf{Answer:} 1.25 / $\tfrac{5}{4}$.
\end{center}

\noindent
\begin{center}
  \begin{minipage}[b]{0.24\textwidth}\centering\includegraphics[width=\linewidth]{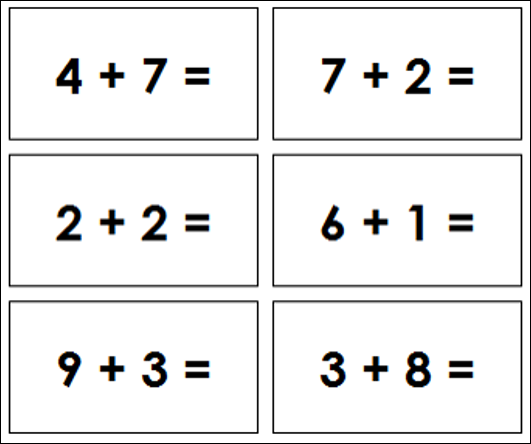}\\(a) Original image\end{minipage}\hfill
  \begin{minipage}[b]{0.24\textwidth}\centering\includegraphics[width=\linewidth]{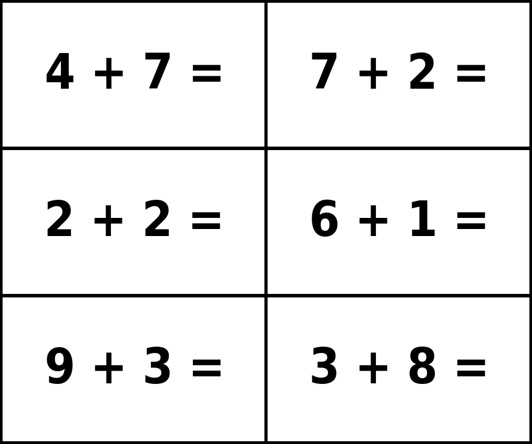}\\(b) Initial rendered\end{minipage}\hfill
  \begin{minipage}[b]{0.24\textwidth}\centering\includegraphics[width=\linewidth]{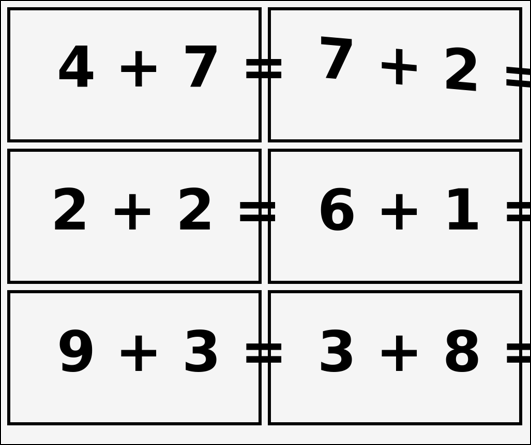}\\(c) w. visual tools\end{minipage}\hfill
  \begin{minipage}[b]{0.24\textwidth}\centering\includegraphics[width=\linewidth]{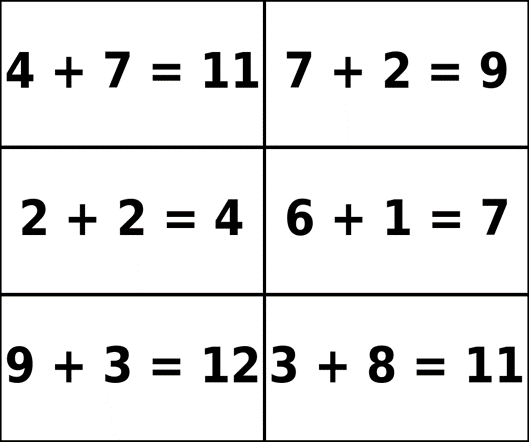}\\(d) w. revision\end{minipage}
\end{center}
\begin{center}
  \textbf{Question:} What is the answer to the second equation on the right? \textbf{Answer:} 12
\end{center}

\noindent
\begin{center}
  \begin{minipage}[b]{0.24\textwidth}\centering\includegraphics[width=\linewidth]{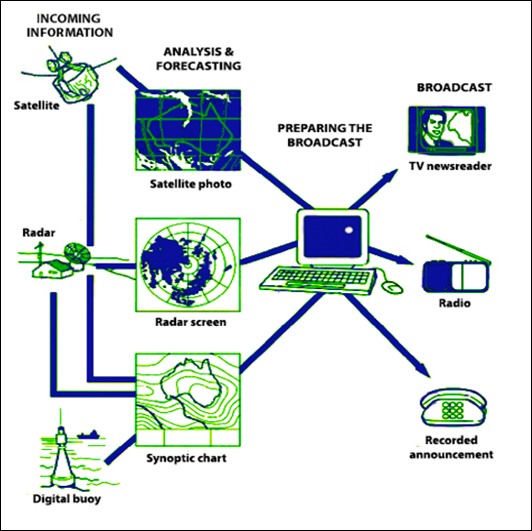}\\(a) Original image\end{minipage}\hfill
  \begin{minipage}[b]{0.24\textwidth}\centering\includegraphics[width=\linewidth]{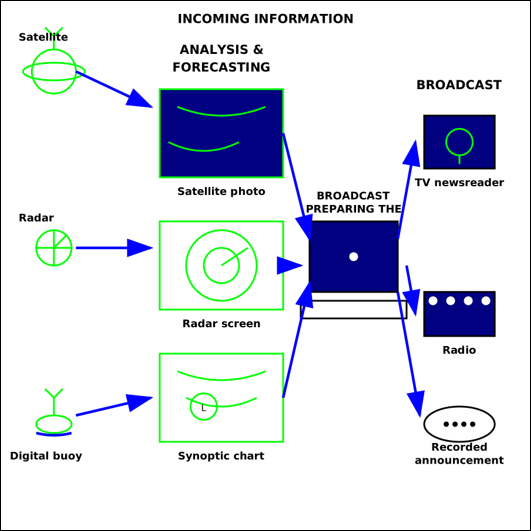}\\(b) Initial rendered\end{minipage}\hfill
  \begin{minipage}[b]{0.24\textwidth}\centering\includegraphics[width=\linewidth]{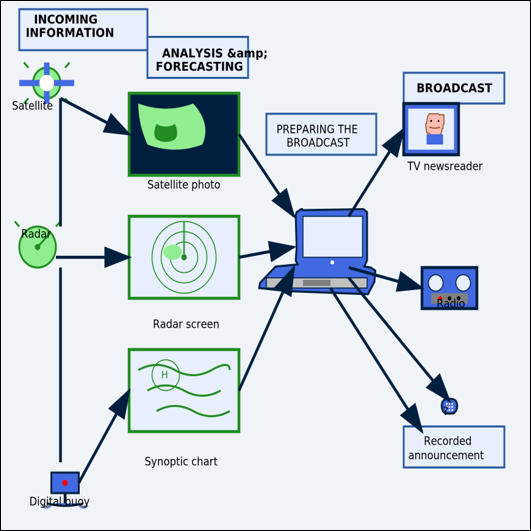}\\(c) w. visual tools\end{minipage}\hfill
  \begin{minipage}[b]{0.24\textwidth}\centering\includegraphics[width=\linewidth]{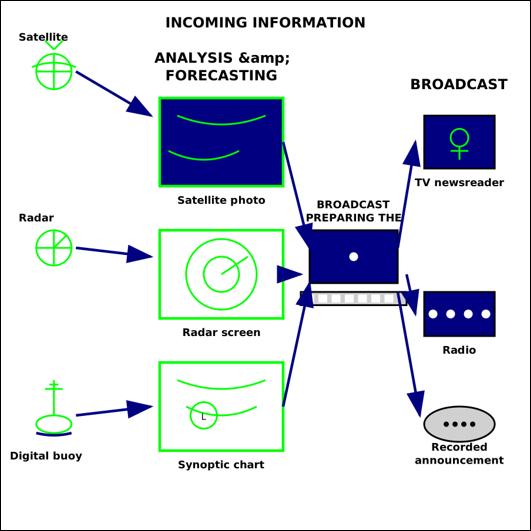}\\(d) w. revision\end{minipage}
\end{center}
\begin{center}
  \textbf{Question:} The diagram below shows how the Australian Bureau of Meteorology collects up-to-the-minute information on the weather in order to produce reliable forecasts. Write a report for a university lecturer describing the information shown below. Write at least 150 words. \\
  \textbf{Answer:} The figure illustrates the process used by the Australian Bureau of Meteorology to forecast the weather. There are four stages in the process, beginning with the collection of information about the weather. This information is then analysed, prepared for presentation, and finally broadcast to the public. Looking at the first and second stages of the process, there are three ways of collecting weather data and three ways of analysing it. Firstly, incoming information can be received by satellite and presented for analysis as a satellite photo. The same data can also be passed to a radar station and presented on a radar screen or synoptic chart. Secondly, incoming information may be collected directly by radar and analysed on a radar screen or synoptic chart. Finally, drifting buoys also receive data which can be shown on a synoptic chart. At the third stage of the process, the weather broadcast is prepared on computers. Finally, it is delivered to the public on television, on the radio, or as a recorded telephone announcement.
\end{center}

\noindent
\begin{center}
  \begin{minipage}[b]{0.24\textwidth}\centering\includegraphics[width=\linewidth]{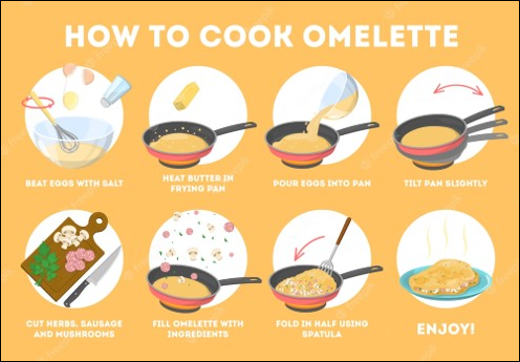}\\(a) Original image\end{minipage}\hfill
  \begin{minipage}[b]{0.24\textwidth}\centering\includegraphics[width=\linewidth]{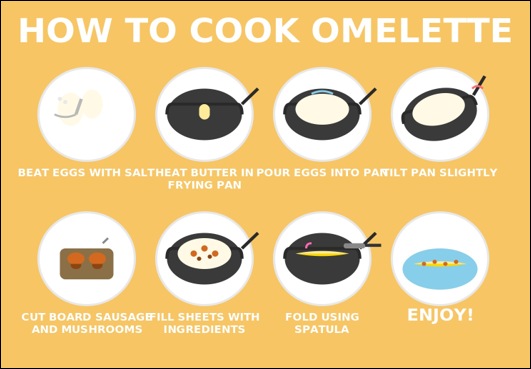}\\(b) Initial rendered\end{minipage}\hfill
  \begin{minipage}[b]{0.24\textwidth}\centering\includegraphics[width=\linewidth]{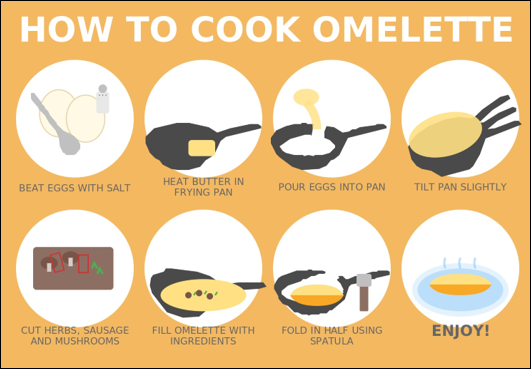}\\(c) w. visual tools\end{minipage}\hfill
  \begin{minipage}[b]{0.24\textwidth}\centering\includegraphics[width=\linewidth]{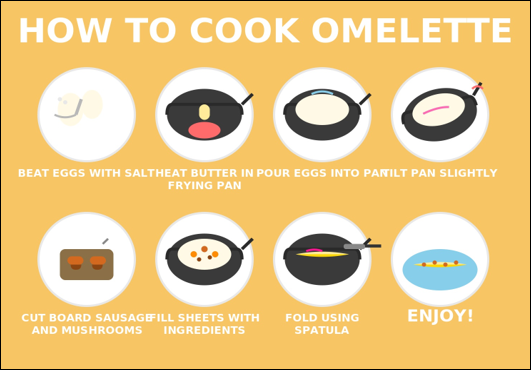}\\(d) w. revision\end{minipage}
\end{center}
\begin{center}
  \textbf{Question:} What should I do before cutting herbs, sausage, and mushrooms? \textbf{Answer:} milk
\end{center}

\noindent
\begin{center}
  \begin{minipage}[b]{0.24\textwidth}\centering\includegraphics[width=\linewidth]{figures/mm-vet/21.png}\\(a) Original image\end{minipage}\hfill
  \begin{minipage}[b]{0.24\textwidth}\centering\includegraphics[width=\linewidth]{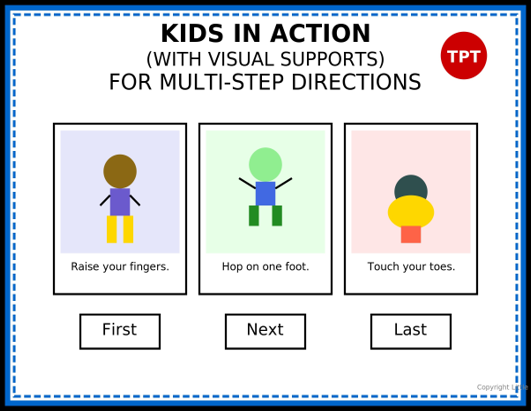}\\(b) Initial rendered\end{minipage}\hfill
  \begin{minipage}[b]{0.24\textwidth}\centering\includegraphics[width=\linewidth]{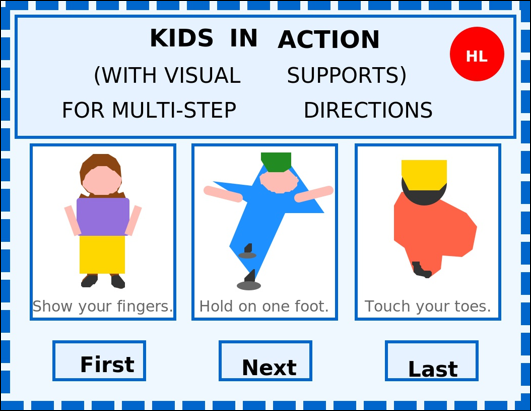}\\(c) w. visual tools\end{minipage}\hfill
  \begin{minipage}[b]{0.24\textwidth}\centering\includegraphics[width=\linewidth]{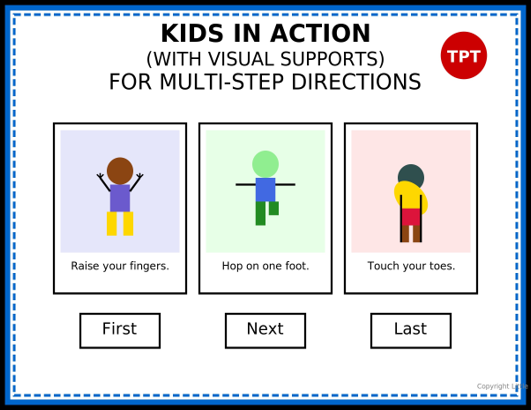}\\(d) w. revision\end{minipage}
\end{center}
\begin{center}
  \textbf{Question:} What should kids do after snap fingers? \textbf{Answer:} hop on one foot
\end{center}

\noindent
\begin{center}
  \begin{minipage}[b]{0.24\textwidth}\centering\includegraphics[width=\linewidth]{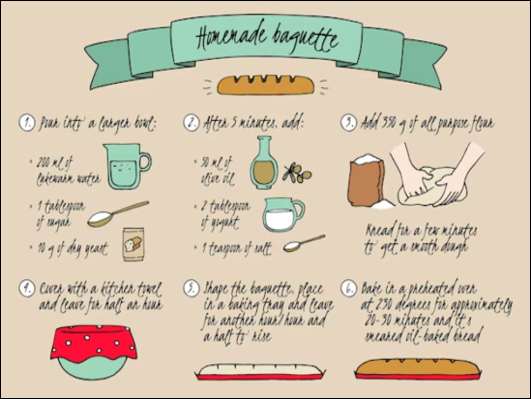}\\(a) Original image\end{minipage}\hfill
  \begin{minipage}[b]{0.24\textwidth}\centering\includegraphics[width=\linewidth]{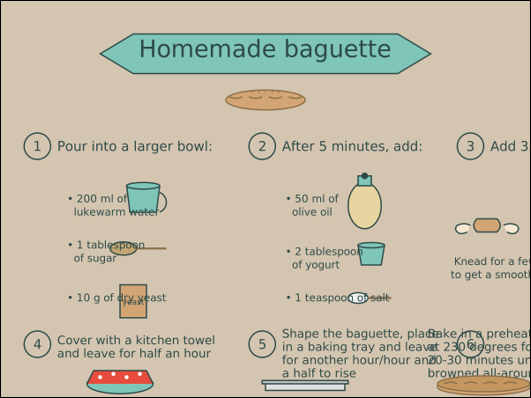}\\(b) Initial rendered\end{minipage}\hfill
  \begin{minipage}[b]{0.24\textwidth}\centering\includegraphics[width=\linewidth]{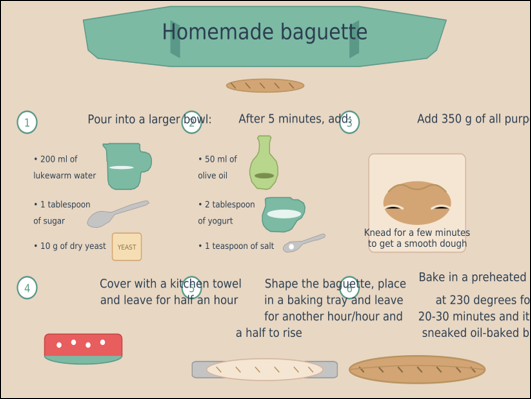}\\(c) w. visual tools\end{minipage}\hfill
  \begin{minipage}[b]{0.24\textwidth}\centering\includegraphics[width=\linewidth]{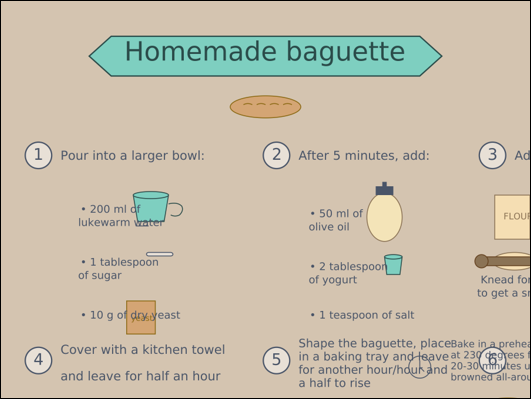}\\(d) w. revision\end{minipage}
\end{center}
\begin{center}
  \textbf{Question:} What is the index of the step when we need to add all purpose flour?
  \textbf{Answer:} third / 3
\end{center}

\noindent
\begin{center}
  \begin{minipage}[b]{0.24\textwidth}\centering\includegraphics[width=\linewidth]{figures/mm-vet/33.png}\\(a) Original image\end{minipage}\hfill
  \begin{minipage}[b]{0.24\textwidth}\centering\includegraphics[width=\linewidth]{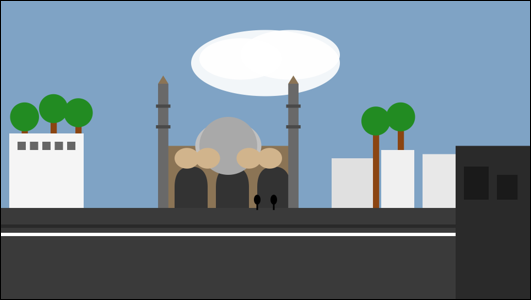}\\(b) Initial rendered\end{minipage}\hfill
  \begin{minipage}[b]{0.24\textwidth}\centering\includegraphics[width=\linewidth]{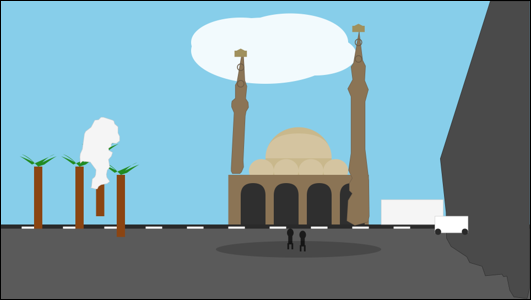}\\(c) w. visual tools\end{minipage}\hfill
  \begin{minipage}[b]{0.24\textwidth}\centering\includegraphics[width=\linewidth]{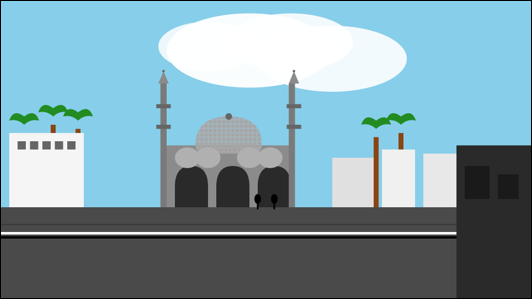}\\(d) w. revision\end{minipage}
\end{center}
\begin{center}
  \textbf{Question:} What is the name of this landmark? \textbf{Answer:} Anbariya Mosque
\end{center}

\noindent
\begin{center}
  \begin{minipage}[b]{0.24\textwidth}\centering\includegraphics[width=\linewidth]{figures/mm-vet/37.png}\\(a) Original image\end{minipage}\hfill
  \begin{minipage}[b]{0.24\textwidth}\centering\includegraphics[width=\linewidth]{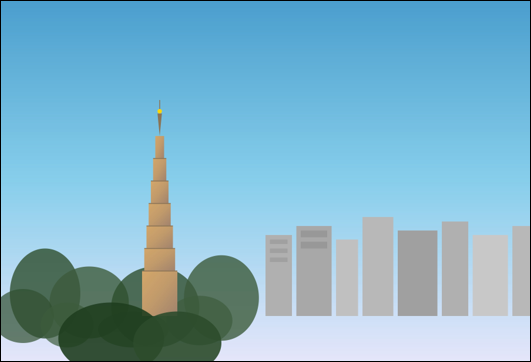}\\(b) Initial rendered\end{minipage}\hfill
  \begin{minipage}[b]{0.24\textwidth}\centering\includegraphics[width=\linewidth]{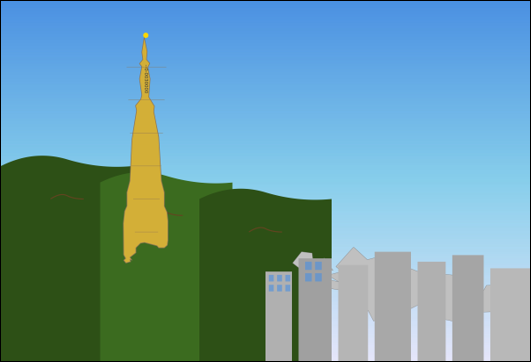}\\(c) w. visual tools\end{minipage}\hfill
  \begin{minipage}[b]{0.24\textwidth}\centering\includegraphics[width=\linewidth]{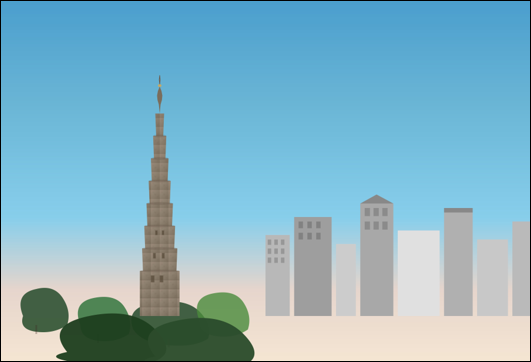}\\(d) w. revision\end{minipage}
\end{center}
\begin{center}
  \textbf{Question:} What is the name of this landmark? \textbf{Answer:} baochu pagoda
\end{center}

\noindent
\begin{center}
  \begin{minipage}[b]{0.24\textwidth}\centering\includegraphics[width=\linewidth]{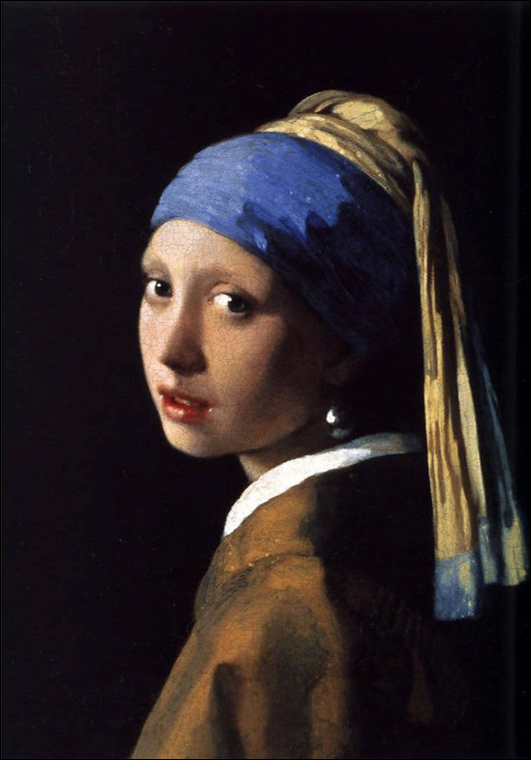}\\(a) Original image\end{minipage}\hfill
  \begin{minipage}[b]{0.24\textwidth}\centering\includegraphics[width=\linewidth]{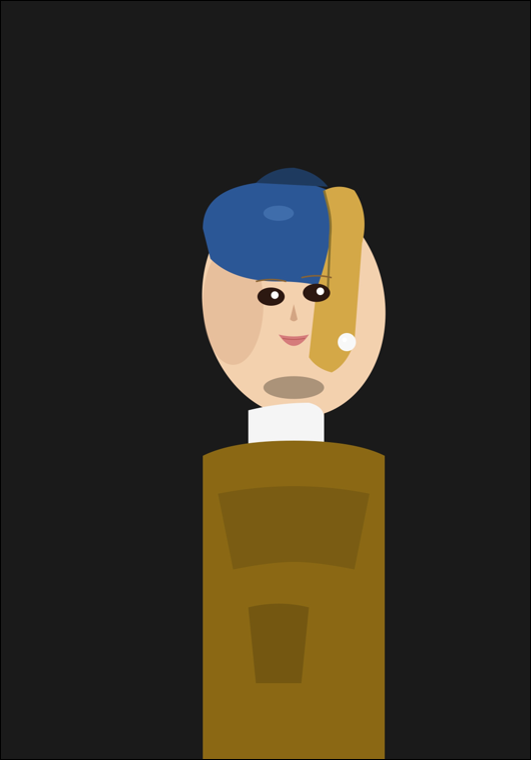}\\(b) Initial rendered\end{minipage}\hfill
  \begin{minipage}[b]{0.24\textwidth}\centering\includegraphics[width=\linewidth]{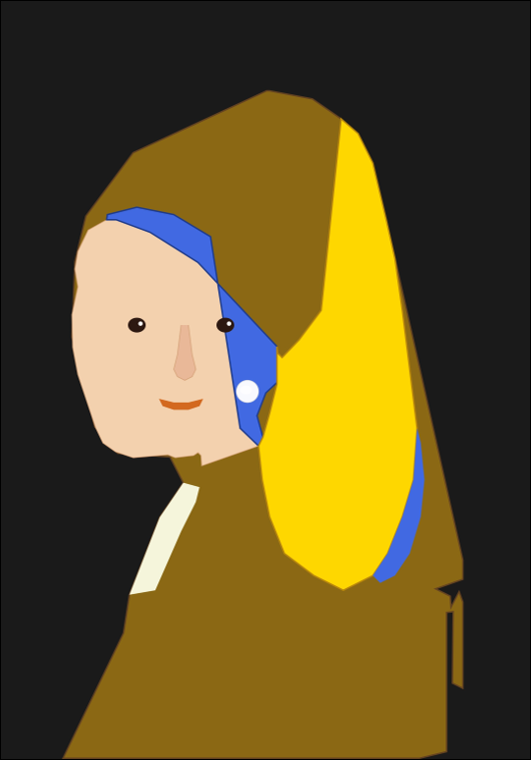}\\(c) w. visual tools\end{minipage}\hfill
  \begin{minipage}[b]{0.24\textwidth}\centering\includegraphics[width=\linewidth]{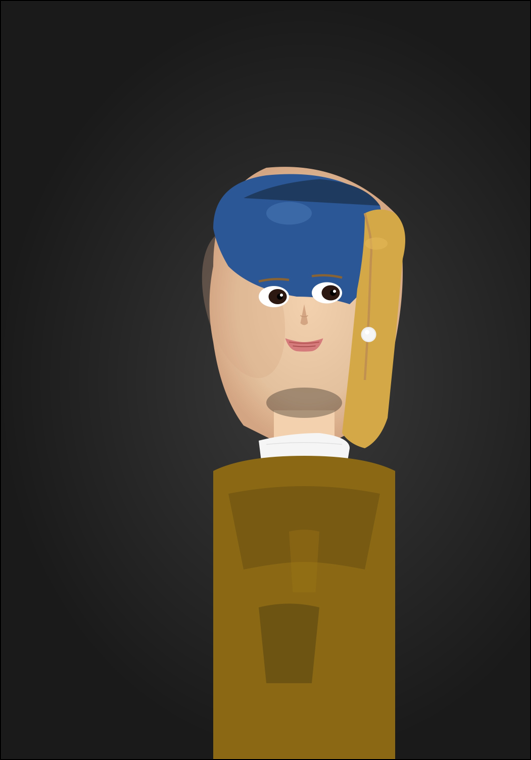}\\(d) w. revision\end{minipage}
\end{center}
\begin{center}
  \textbf{Question:} Can you give a short introduction to this painting? \\
  \textbf{Answer:} Girl With A Pearl Earring (Dutch: Meisje met de parel) is an oil painting by Dutch Golden Age painter Johannes Vermeer, dated c. 1665. Going by various names over the centuries, it became known by its present title towards the end of the 20th century after the earring worn by the girl portrayed there. The work has been in the collection of the Mauritshuis in The Hague since 1902 and has been the subject of various literary and cinematic treatments..
\end{center}

\noindent
\begin{center}
  \begin{minipage}[b]{0.24\textwidth}\centering\includegraphics[width=\linewidth]{figures/mm-vet/49.png}\\(a) Original image\end{minipage}\hfill
  \begin{minipage}[b]{0.24\textwidth}\centering\includegraphics[width=\linewidth]{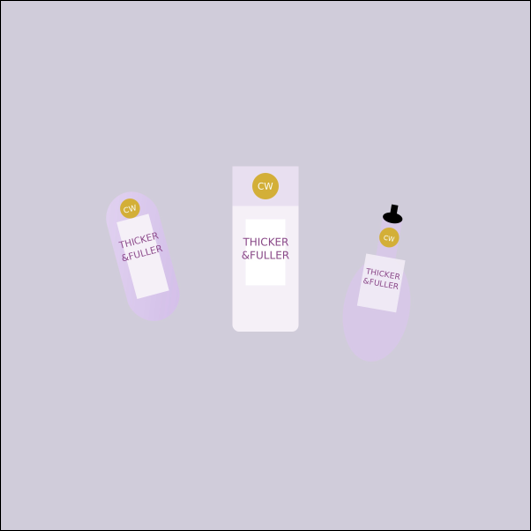}\\(b) Initial rendered\end{minipage}\hfill
  \begin{minipage}[b]{0.24\textwidth}\centering\includegraphics[width=\linewidth]{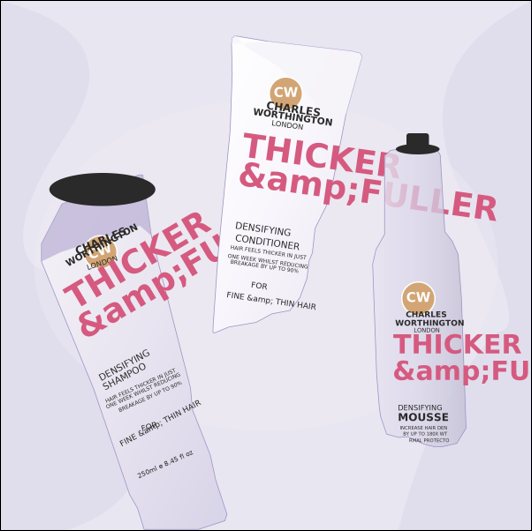}\\(c) w. visual tools\end{minipage}\hfill
  \begin{minipage}[b]{0.24\textwidth}\centering\includegraphics[width=\linewidth]{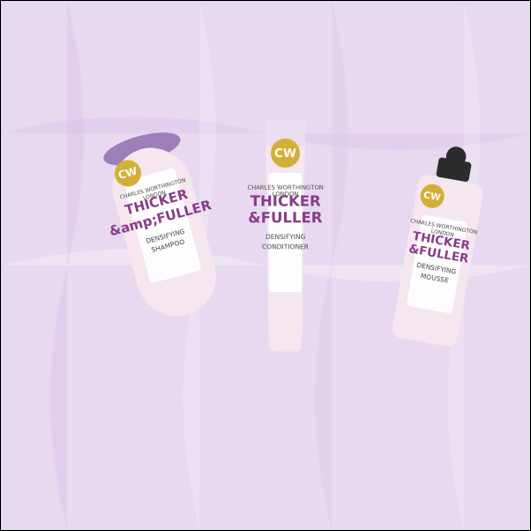}\\(d) w. revision\end{minipage}
\end{center}
\begin{center}
  \textbf{Question:} What is located to the right of the shampoo? \textbf{Answer:} conditioner
\end{center}

\noindent
\begin{center}
  \begin{minipage}[b]{0.24\textwidth}\centering\includegraphics[width=\linewidth]{figures/mm-vet/53.png}\\(a) Original image\end{minipage}\hfill
  \begin{minipage}[b]{0.24\textwidth}\centering\includegraphics[width=\linewidth]{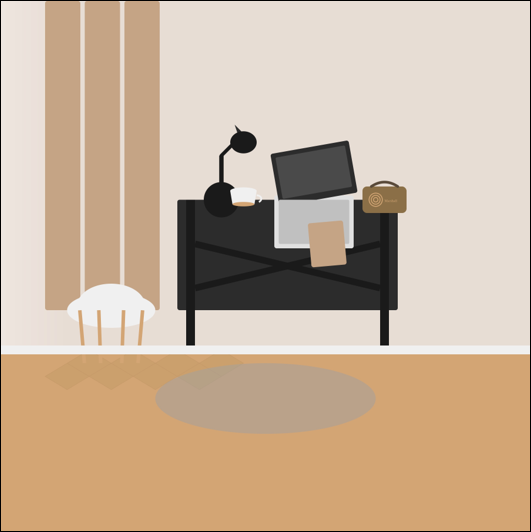}\\(b) Initial rendered\end{minipage}\hfill
  \begin{minipage}[b]{0.24\textwidth}\centering\includegraphics[width=\linewidth]{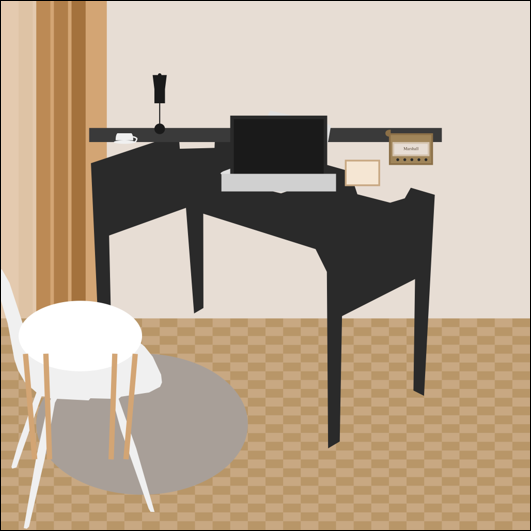}\\(c) w. visual tools\end{minipage}\hfill
  \begin{minipage}[b]{0.24\textwidth}\centering\includegraphics[width=\linewidth]{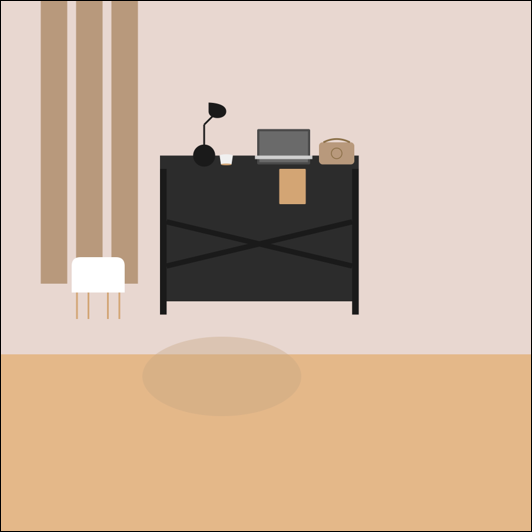}\\(d) w. revision\end{minipage}
\end{center}
\begin{center}
  \textbf{Question:} Is the curtain on the right side or on the left of the picture? \textbf{Answer:} left
\end{center}

\noindent
\begin{center}
  \begin{minipage}[b]{0.24\textwidth}\centering\includegraphics[width=\linewidth]{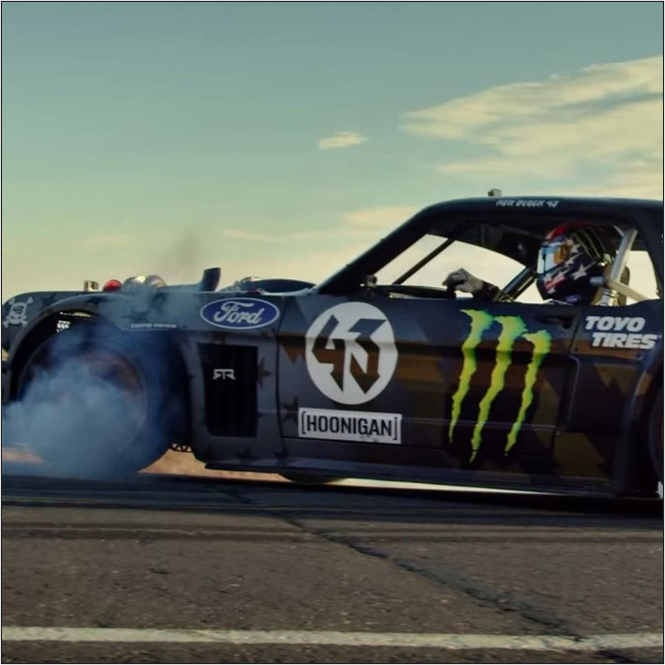}\\(a) Original image\end{minipage}\hfill
  \begin{minipage}[b]{0.24\textwidth}\centering\includegraphics[width=\linewidth]{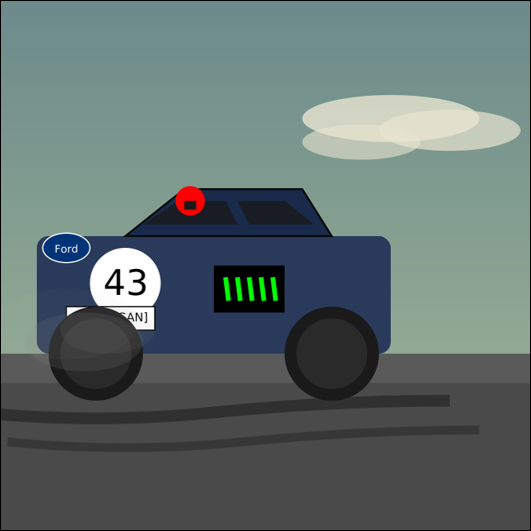}\\(b) Initial rendered\end{minipage}\hfill
  \begin{minipage}[b]{0.24\textwidth}\centering\includegraphics[width=\linewidth]{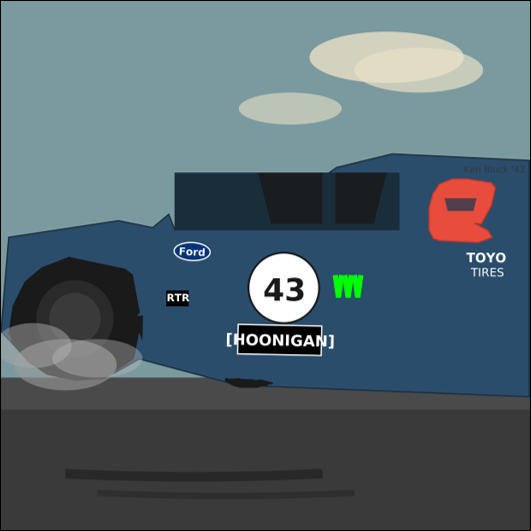}\\(c) w. visual tools\end{minipage}\hfill
  \begin{minipage}[b]{0.24\textwidth}\centering\includegraphics[width=\linewidth]{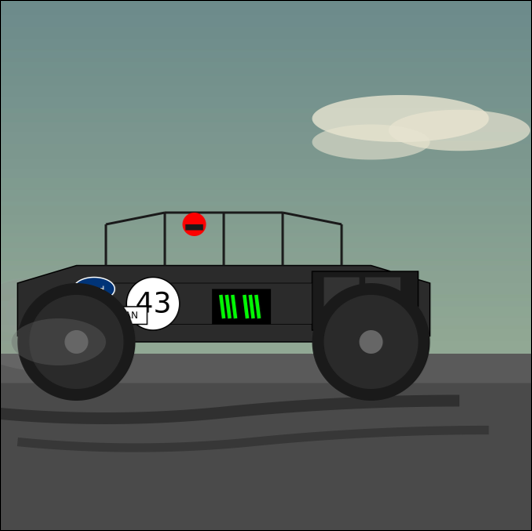}\\(d) w. revision\end{minipage}
\end{center}
\begin{center}
  \textbf{Question:} what is the green logo on the car? \textbf{Answer:} monster.
\end{center}

\clearpage

\end{document}